# FaCells—Teaching Machines the Language of Lines: Per-Point Attribute Scores for Face-Sketch Classification

Xavier Ignacio González
School of Engineering
University of Buenos Aires
Ciudad de Buenos Aires, Argentina
xgonzalez@fi.uba.ar

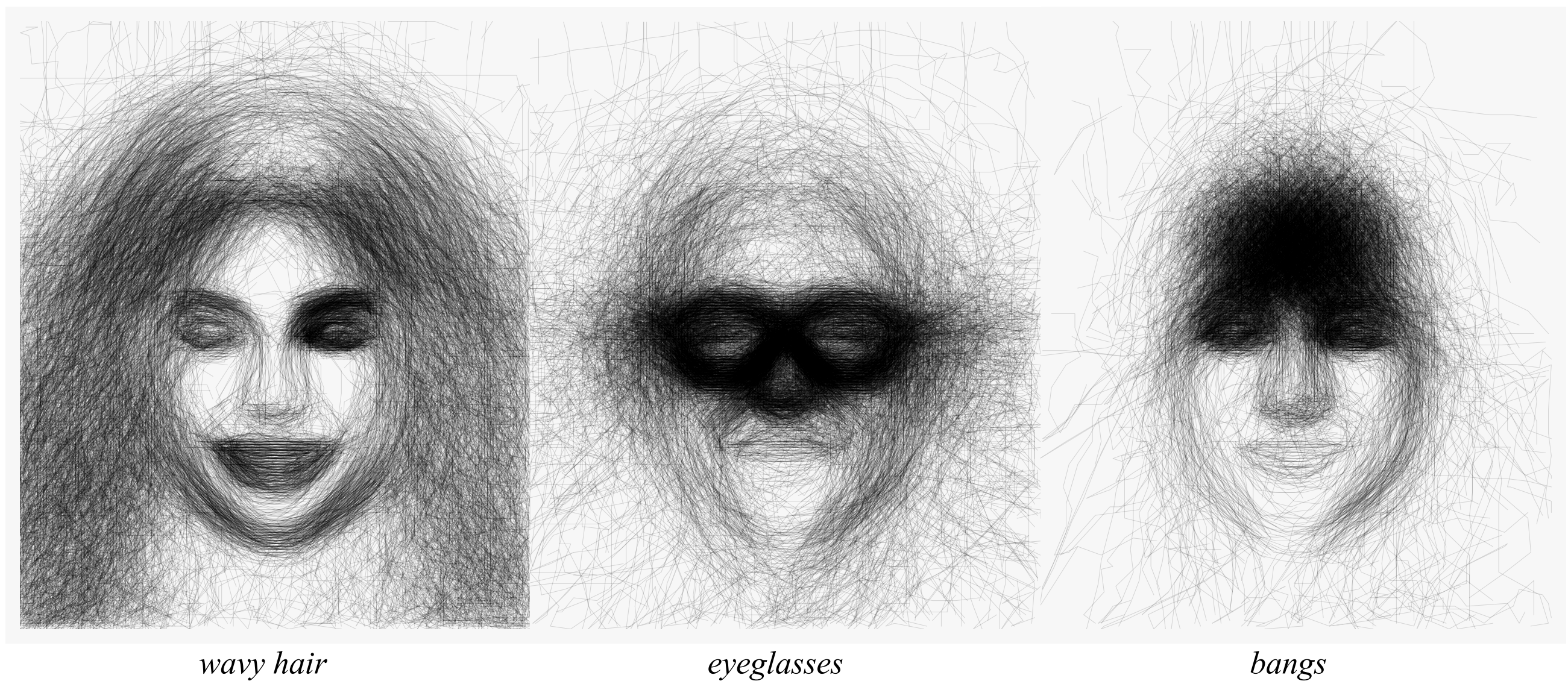

*Figure 1. FaCells for three attributes—wavy hair, eyeglasses, and bangs—created by overlaying 1,000 vector face sketches. For each sketch, a modified LSTM assigns a per-point attribute score; points above a threshold are retained and composited into the final image.*

***Abstract*— FaCells is a method—and an exhibition—that turns model internals into line-based artworks. Aligned face photographs (CelebA, 260k images, 40 attributes) are translated into vector sketches suitable for an XY plotter. We study how to “write” these drawings for a sequence model, comparing absolute vs. relative point encodings and random vs. travel-minimizing stroke order. A bidirectional LSTM is trained for attribute prediction; a minimal architectural change—removing the global average over the sequence and applying a Dense layer at each point—yields per-point attribute scores. Aggregating points whose score exceeds an attribute-specific threshold across many portraits produces new drawings we call FaCells: statistical abstractions of attributes such as Eyeglasses, Wavy Hair, or Bangs.**

**Across ablations, absolute coordinates with travel-minimizing order and a global-average readout perform best; this configuration is then adapted to produce per-point scores. Multilabel training over 40 attributes is stable, and attributes reaching at least 50% balanced accuracy are visualized as FaCells. Complementary notions (e.g., No_Beard) are constructed by selecting points below a negative threshold. FaCells foregrounds interpretability as a creative tool: the resulting works are plotter-ready, reproducible, and inexpensive to realize, yet materially present. Presented at Spectrum Miami 2025, the project bridges data, model, and paper while acknowledging the limits of the labels and the biases of the dataset.**

## I. Introduction

“A line is a dot that went for a walk,” wrote Paul Klee. In this project, that walk becomes both an artistic gesture and a computation. A line is not only a trace of a hand on paper, but also a sequence of coordinates, a path a machine can read, transform, and reinterpret. FaCells emerges from that dual nature: it is an attempt to understand what a learned machine “sees” in thousands of face drawings, and to turn that internal view back into new line-based artworks.

Line drawing is one of the oldest human abstractions. Long before photography or cinema, a few strokes were enough to suggest a body, a face, an animal, an idea. In contemporary visual culture, however, photographic and video images have largely replaced line drawing as a primary way of recording reality. Yet lines never disappeared; they persist in sketches, comics, vector graphics, and plotter drawings, and they remain an efficient way to compress visual information into something that is at once simple and expressive. In a digital context, a line can be described as a list of points and instructions—"move here, move there, pen up, pen down"—lightweight but capable of evoking complex forms in the viewer's mind.

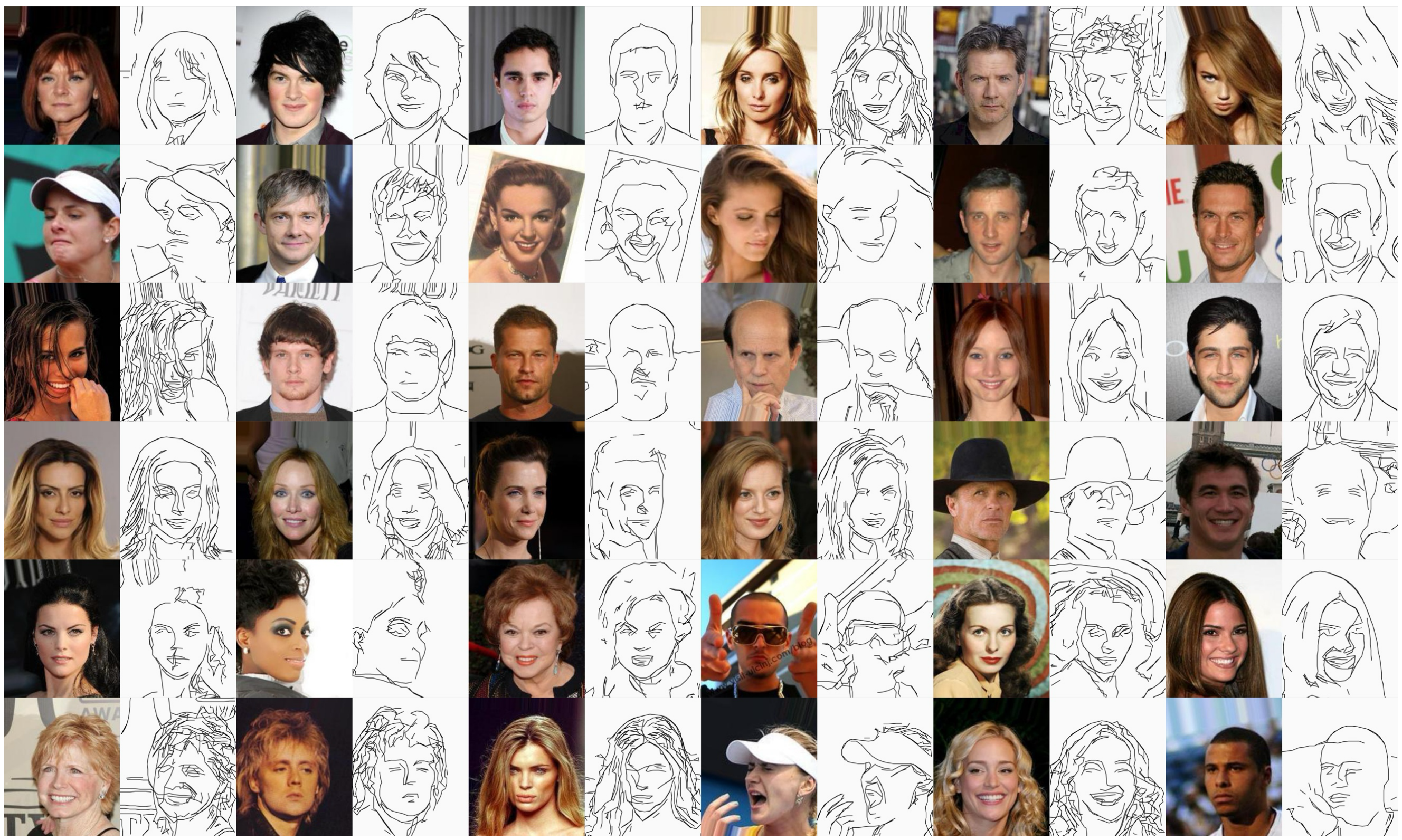

*Figure 2. CelebA photographs paired with their sketch-based portraits, generated through an iterative contour-to-vector reconstruction.*

The images in this work begin as portraits from the CelebA dataset, a collection of more than 260,000 aligned face photographs annotated with 40 visual attributes [1]. Each photograph is translated into a vector sketch using an iterative edge-based procedure: contours are detected and then redrawn as polylines suitable for an XY plotter or any other device that can follow line instructions. A portrait, in this representation, is no longer pixels but a set of lines, and each line is no more than a sequence of points. Figure 2 shows example pairs of original faces and their derived sketch-based portraits.

On top of this representation, a bidirectional LSTM network is trained on the portraits with the stated task of predicting the dataset's binary "Male" attribute. Training begins with this single attribute and then extends to all 40 attributes in [1]. This target is chosen not because gender classification is a meaningful or desirable end in itself, but because it allows us to shift the focus away from performance and toward representation and interpretation. We note that the dataset label is binary, whereas gender in the world is not. The goal is not to build the best classifier, but to observe how the model organizes the drawings internally, and then to reuse that organization creatively.

A key idea of FaCells is to look at the model not only as a device that produces a single score per portrait, but as a device that can assign a score to each point of each drawing: a measure of how strongly that point contributes to an attribute such as "eyeglasses" or "smiling." By modifying the network architecture to output per-point scores, we can select a large collection of sketches and keep only the points whose score passes a threshold. If we overlay those selected points from many different faces, the result is a new drawing: a distilled, statistical abstraction of an attribute. These aggregated drawings are what this work calls FaCells.

From an artistic perspective, the FaCells are proposed as an "objective" visual essence of an attribute—objective in the limited sense that no individual human style intervenes in their construction. They are not someone's drawing of eyeglasses or a smile; they are what remains when many labeled examples are projected through a trained model and only the most characteristic points survive. From a technical perspective, they can be seen as a form of model interpretability reimagined as an aesthetic object: instead of heatmaps over pixels, they are line drawings over a shared canvas.

Presented in the context of the Spectrum Miami 2025 exhibition[1], this paper documents the data preparation and model design with enough detail for reproducibility while reflecting on the aesthetic and conceptual implications of FaCells. It treats the plotter as a medium that bridges the digital and the physical: code drives motion, vectors become marks on paper, and per-point model scores are encountered as lines. This framing—between code and plotter, between attribute and line—invites readers and viewers to consider how machine learning, labels, and large datasets can generate not only predictions, but also new visual languages using lines.

Following this introduction, Section II reviews related work and sources of inspiration. Section III outlines the method—from photos to vector sketches, representation and line ordering, model variants, and the small architectural tweak that yields per-point attribute scores. Section IV reports results from the three training stages. Section V presents the FaCells, how they are assembled, and how to read their titles. Section VI frames the Spectrum Miami 2025 exhibition and the plotter as a bridge from code to paper. Section VII discusses interpretation, limitations, and future directions. The paper concludes with acknowledgments and references.

## II. Related Work and Sources of Inspiration

This project builds on three threads. First, robotic drawing systems such as Portrait Drawing by Paul the Robot translate photographs into plotter-ready strokes without explicit facial-part models, foregrounding the plotter as a medium and vector strokes as the representational unit [2]. Second, sequence models show that networks can learn stroke dynamics: an LSTM can synthesize realistic handwriting with pen-up/pen-down states, and sketch-RNN models object drawings as probabilistic stroke sequences [3], [4]. Third, photo-to-sketch translation has been studied both broadly and for portraits, from edge- and contour-based stylization to analyses of how facial structure is abstracted in sketching [5]–[7]; the reverse direction—sketch-to-photo synthesis—has likewise been explored [8]. Our data and supervision come from CelebA's aligned faces and binary attributes [1], with standard deep-learning methodology as background [9]. FaCells departs from prior art by treating vector sketches as inputs to a bidirectional LSTM trained on attributes, modifying the architecture to output per-point scores, and aggregating those points across many drawings to produce new line-based compositions rather than optimizing for generation or classification alone.

## III. Methods and Experiments

We begin by turning photographs into lines that a plotter can draw, then decide how those lines should be written down for a sequence model to "read." With that in place, we train a family of bidirectional LSTMs inspired by prior sketch-generation work and make one small architectural change that lets the network assign a score to every point in a drawing—the raw material of the FaCells.

### *1) From photograph to sequence (vector lines)*

Starting from the aligned CelebA faces [1], each portrait is translated into a vector sketch: contours are detected, traced as polylines. We preserve pen state explicitly: points belong to strokes when the pen is down, and stroke boundaries are marked when the pen lifts. This yields, for every portrait, a compact sequence of 2D points grouped into strokes that the plotter can execute.

### *2) How we "write" a drawing: encoding and ordering*

Because the model consumes sequences, how we encode a point matters. We therefore compare two natural choices. In the relative (Δ) format, adapted from sketch-RNN and earlier handwriting models [4], [3], each point is $(\Delta x, \Delta y, p)$, where $\Delta x$ and $\Delta y$ are offsets from the previous point and $p$ encodes pen state (start / continue / end, implemented as $\{1, 0, -1\}$). This description is compact and intentionally insensitive to where the drawing sits on the page. In the absolute format, each point is $(x, y, p)$ in a shared canvas. Because CelebA faces are aligned and scale-normalized, absolute coordinates carry useful global layout cues (eyes, mouth, hair tend to live in similar places), which a purely relative code erases.

Ordering the strokes also shapes what the network sees. For each sketch we create two instances: unsorted (random stroke order) and sorted, where strokes are arranged to minimize total travel, including pen-up hops, of an idealized plotter. We reduce the problem to a Traveling-Salesman-like routing over stroke endpoints and solve it with Google OR-Tools heuristics [10]. Although the routing is NP-hard and must be solved >200K times (with roughly 100–1000 strokes per sketch), OR-Tools makes the preprocessing practical. Intuitively, keeping strokes from the same facial element (e.g., the two brows, the chin contour) nearby in time simplifies the temporal structure the LSTM must learn.

### *3) Model, per-point scores, and training*

Our models follow the encoder half of sketch-RNN [4]: bidirectional LSTMs that read the point sequences and output attribute predictions. We explore a small family rather than chase a single "best" classifier.

- **Depth.** 1bi has one BiLSTM (256 units); 3bi stacks three BiLSTMs (150 units each).
- **Readout.** fs ("final state") feeds the last hidden state to a Dense classifier; ga ("global average") keeps the full sequence, applies GlobalAveragePooling1D over time, then a Dense output.
- **Head size.** −d1 uses a direct output layer; −d40 inserts a Dense(40) hidden layer before the output (sigmoid).

A single, crucial tweak turns these classifiers into point-wise scorers. Starting from the better-behaved ga variant, we remove the GlobalAveragePooling1D layer and replace it with a Dense layer applied at every sequence position—that is, a per-point prediction head (e.g., Dense(15) for the 15 attributes we visualize, or Dense(40) for the full set). In other words, instead of averaging all hidden states and predicting once, the network now predicts at each point along the line: per-point scores that indicate how strongly a given point contributes to an attribute. Practically, we initialize this head from the trained classifier and use the resulting scores as relative evidence for later selection and aggregation (Section V).

[1] https://redwoodartgroup.com/spectrum-miami/

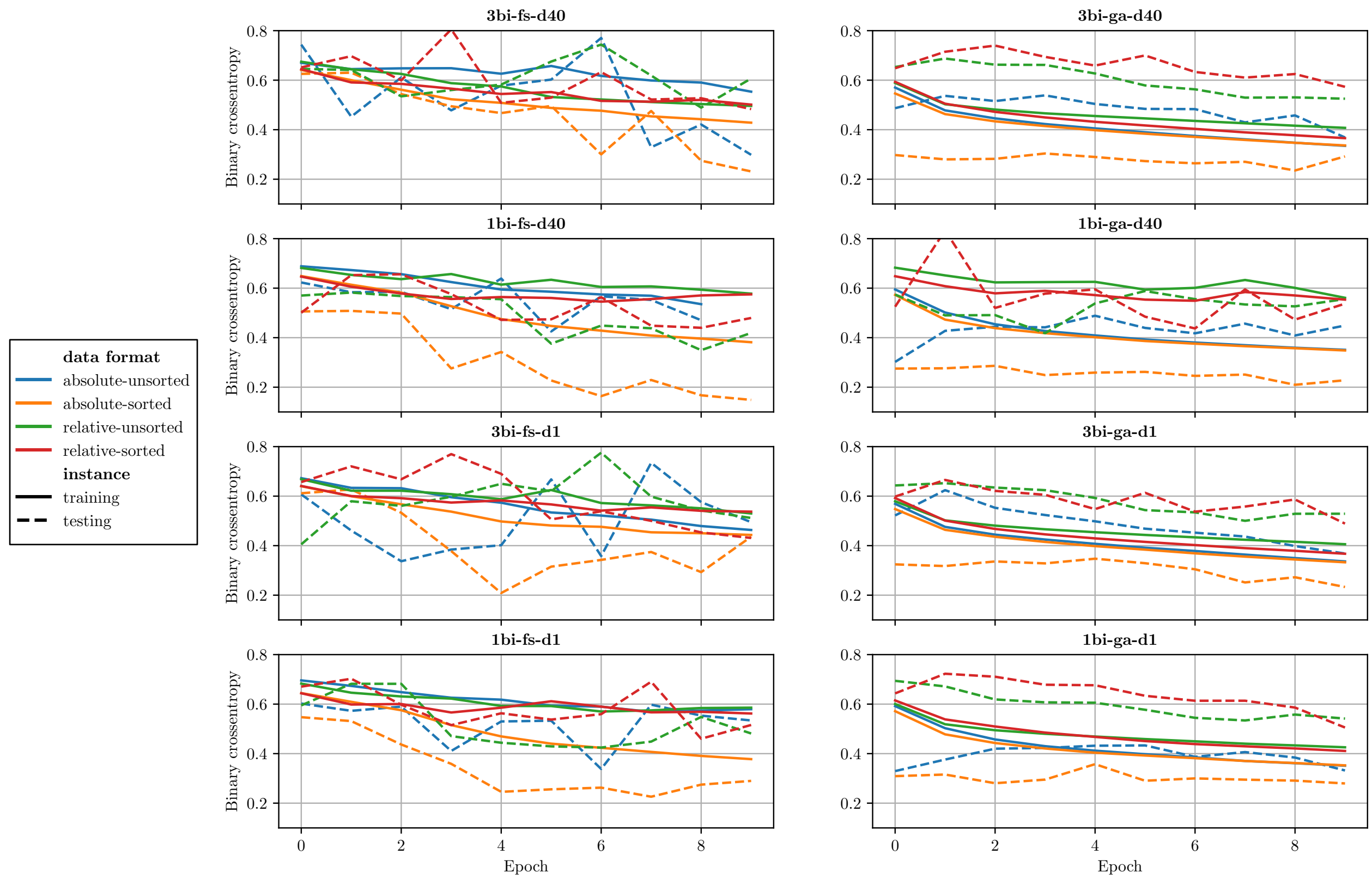

*Figure 3. Stage I of training. Each sub-chart shows a particular LSTM configuration model training for proposed data formats, represented by colored lines. The configuration labels are 1bi (3bi), one LSTM layer with 256 cells (a series of 3 LSTM layers with 150 cells each); d1 (d40), one cell fully-connected output layer (one fully-connected layer with 40 cells, following with one cell fully-connected output layer).*

Training uses consistent hyperparameters across all experiments: batch size 16, the Adam optimizer with default Keras settings (learning rate 0.001), and no explicit regularization. All runs were stable, with no anomalies observed. Performance is monitored using binary cross-entropy during training and balanced accuracy for attribute evaluation. Attributes that reach at least 50% balanced accuracy are considered for visualization as FaCells.

## IV. Results

We now report the outcomes of the three experimental stages. Stage I compares different ways of encoding and ordering the sketches, and evaluates the two readout strategies. Stage II trains the strongest candidates at full scale. Stage III extends the selected configuration to multilabel prediction across forty attributes and determines which attributes are reliable enough to visualize as FaCells.

### A. Stage I — What the model “likes” to read

Using a training subset of $\approx$30% of the drawings (~60K) and a held-out test set of $\approx$ 15%, we ran 10 epochs and compared binary cross-entropy (see Fig. 3). Two patterns emerged clearly:

- *How you write the drawing matters.* The absolute + sorted instance (absolute coordinates with travel-minimizing stroke order) outperformed the other three combinations, which clustered together at a lower level. Intuitively, aligned faces give absolute coordinates a stable spatial anchor, and minimizing pen-up travel keeps related facial strokes close in time.
- *How you read the sequence matters.* Models with global-average readout (ga) consistently beat those that used only the final state (fs).

Among configurations based on the absolute-sorted input, the top three after 10 epochs were 3bi-ga-d1 (0.3328), 3bi-ga-d40 (0.3348), and 1bi-ga-d40 (0.3484) in binary cross-entropy. These finalists advanced to Stage II.

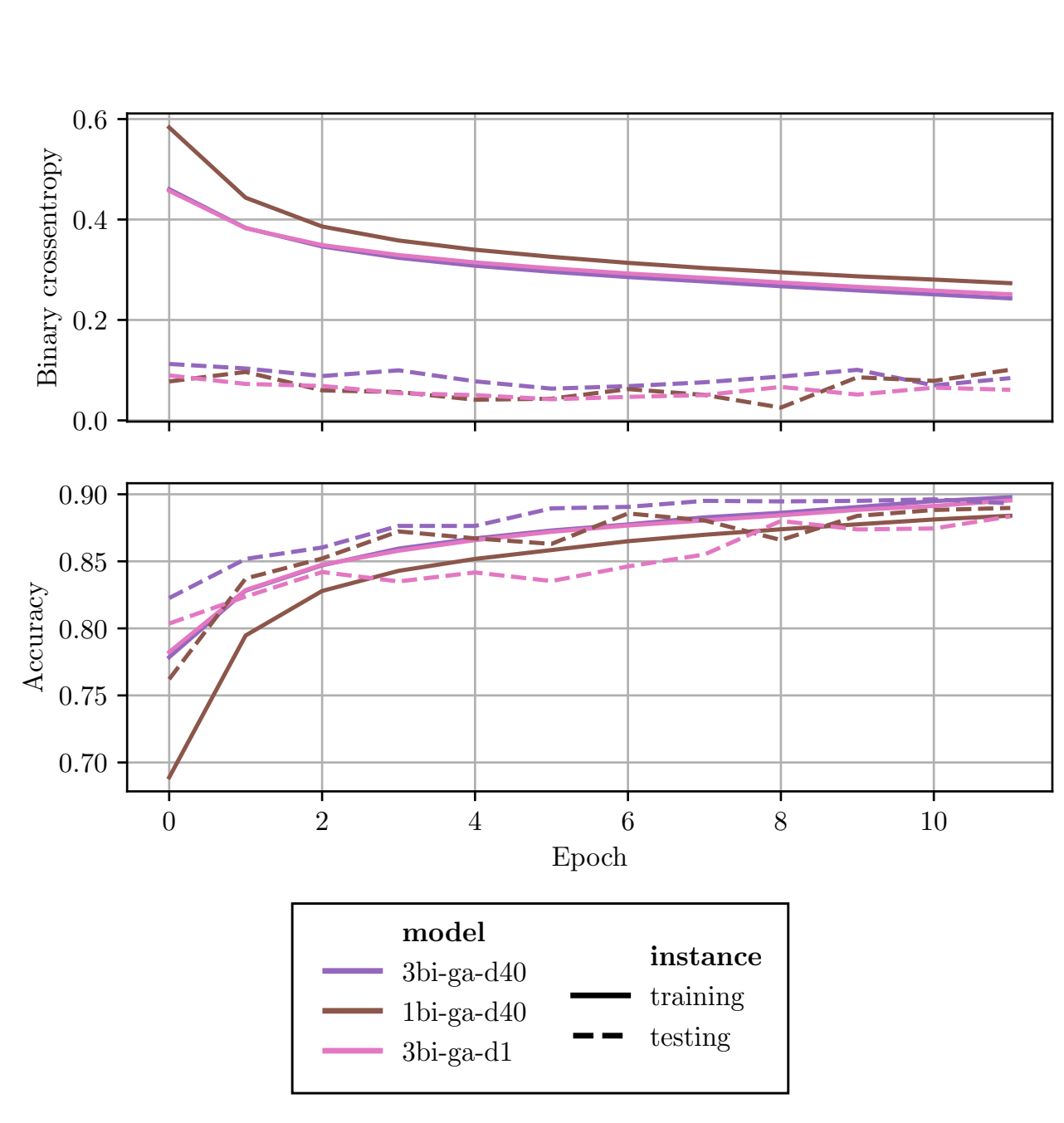

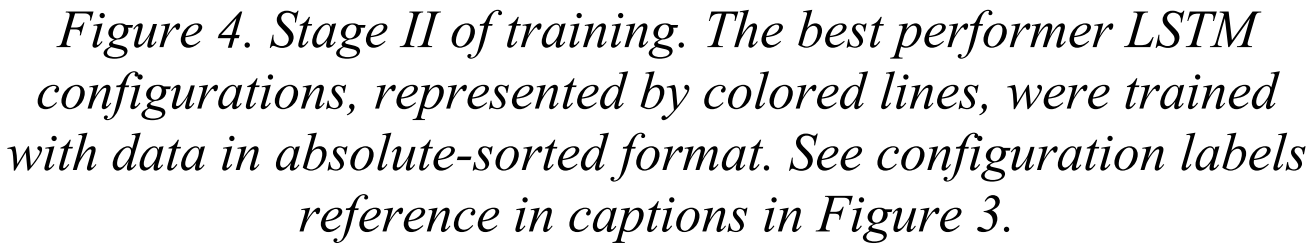

Figure 4. *Stage II of training. The best performer LSTM configurations, represented by colored lines, were trained with data in absolute-sorted format. See configuration labels reference in captions in Figure 3.*

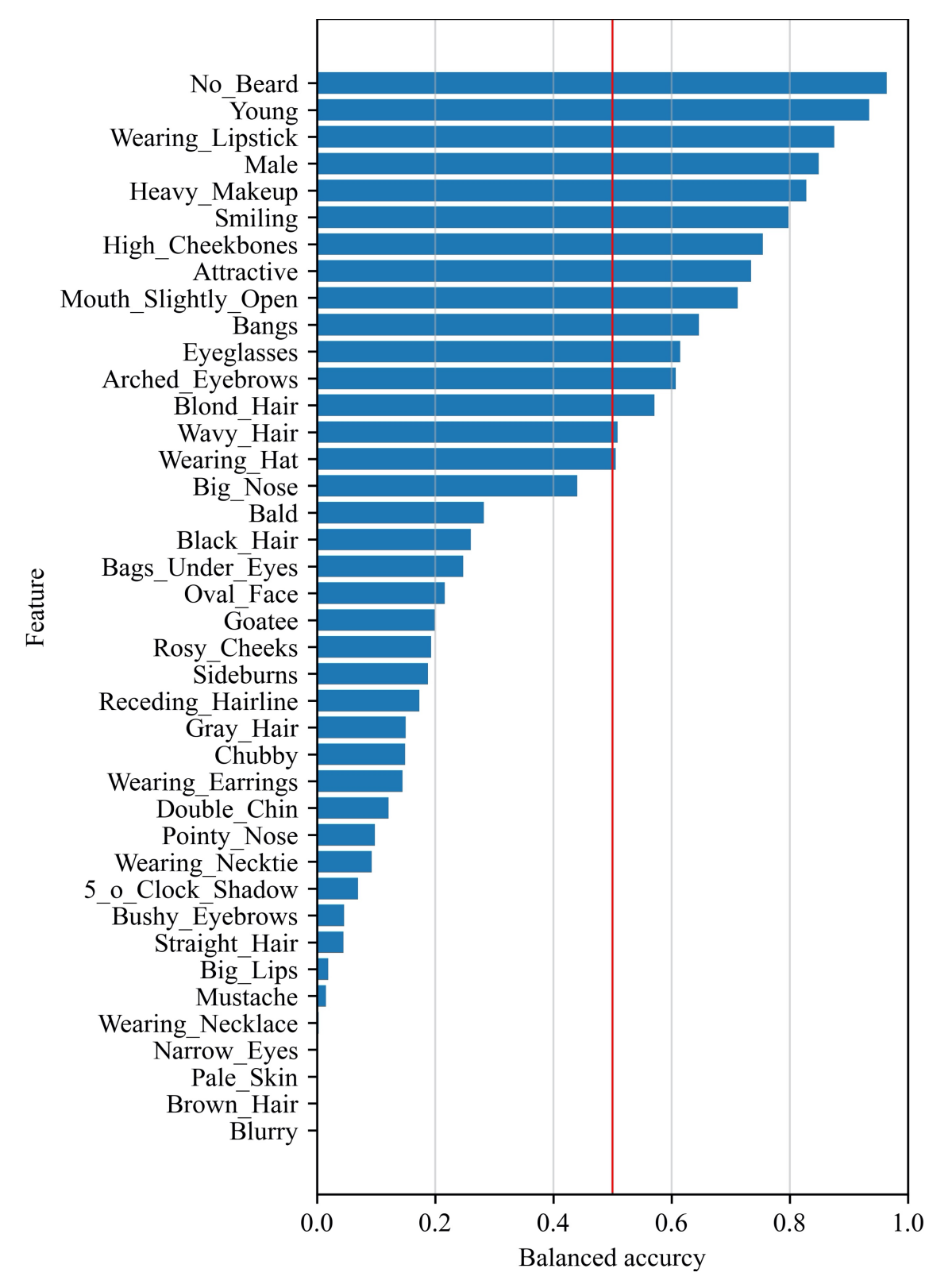

Figure 5. *Stage III of training. Report of balanced accuracy by attribute in testing instance of an LSTM configured as 3bi-ga-d40, the last dense layer is also the output, and the input is in absolute-sorted format.*

### *B. Stage II — Scaling up without surprises*

We trained the three finalists on the full dataset (~260K drawings; ~95/5 train/test) for another 10 epochs (Fig. 4). Performance remained close, with 3bi-ga-d1 showing a slight advantage in the final epochs' cross-entropy. Given its simplicity at the head and stable behavior, we selected 3bi-ga-d1 as the base model moving forward.

### *C. Stage III — Multilabel performance and a cut line for FaCells*

We adapted the winning configuration for multilabel classification over 40 attributes and retrained for 10 epochs on all drawings. Figure 5 reports balanced accuracy per attribute and marks 50% with a vertical red line as the inclusion threshold. Attributes meeting or exceeding this level were deemed reliable enough to visualize as FaCells. (Balanced accuracy is appropriate here because it gives equal weight to positive and negative cases, mitigating class imbalance.)

### *D. Takeaways*

Across stages, three design choices made a consistent difference: (1) absolute coordinates over relative ones when drawings are aligned; (2) travel-minimizing stroke order rather than random order; and (3) a global-average sequence readout (ga) over a final-state readout (fs). Together, these choices produced a model whose internal sequence signals are strong and coherent—ideal conditions for the small architectural tweak that yields per-point scores. In the next section we turn those scores into images by filtering and overlaying points across many portraits—the FaCells.

## V. The FaCells

FaCells begin once the model produces an attribute score for every point in a drawing. Starting from the configuration selected in Section IV, we use the per-point Dense head described in Section III to obtain scores for each attribute. For every portrait, this yields a sequence of per-point scores aligned with the spatial progression of the lines—scores indicating how strongly each point contributes according to the trained network.

Each FaCell is directly tied to an output cell of the Dense layer and to its CelebA attribute. For example, FaCell 4 uses the per-point scores from Dense output cell 4, which corresponds to Eyeglasses in the attribute index; the visualization is constructed from those scores.

To assemble a FaCell for a chosen attribute, points from many portraits are collected whenever their per-point score at the Dense output exceeds a threshold. The threshold is not fixed; it varies by attribute and by drawing. Higher values isolate a cleaner structural core, while lower values include a broader, more diffuse expression. Thresholds can also be negative, allowing complementary constructions such as No_Male, No_Young, No_Attractive, or Beard (derived from the CelebA label No_Beard by selecting points below a negative threshold). The same procedure applies symmetrically in both directions. Attributes that meet the inclusion criterion described in Section IV (balanced accuracy $\geq 50\%$) are considered for visualization.

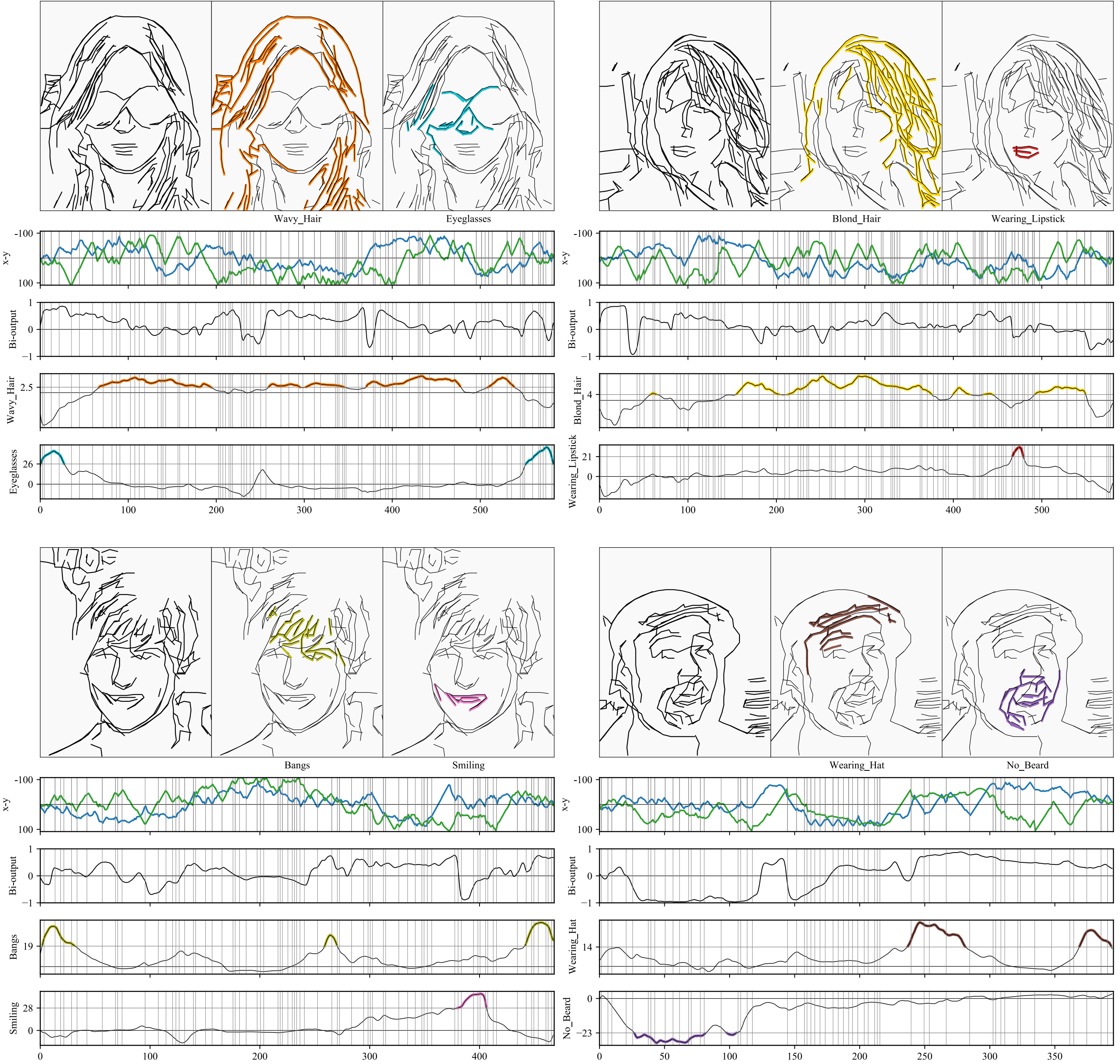

*Figure 6. Four portraits with two attribute annotations each: Wavy_Hair, Eyeglasses, Blond_Hair, Wearing_Lipstick, Bangs, Smiling, Wearing_Hat, and Beard (from No_Beard). For each portrait, the plot labeled "x–y" unfolds the drawing along its sequence, with the x- and y-coordinates shown as alternating blue and green traces; thin vertical gray lines mark stroke boundaries. The plot labeled "Bi-output" shows the sequence returned by the first unit of the third LSTM layer. The two plots labeled with attribute names display the corresponding per-point outputs of the modified head. A threshold selects the points for each attribute; the same colors are used on the portrait to indicate the associated line segments.*

Figure 6 illustrates four portraits with two attributes each. In the examples shown, attributes include Wavy_Hair, Eyeglasses, Blond_Hair, Wearing_Lipstick, Bangs, Smiling, Wearing_Hat, and No_Beard. Points that satisfy the threshold are highlighted and echoed on the portrait in matching colors. Aggregating such points over many portraits produces the FaCells presented in Figure 1 and Appendix I.

## VI. Final Comments

At Spectrum Miami 2025, FaCells is presented as a passage from dataset to drawing: code organizes thousands of portraits into sequences of lines, and a plotter inscribes those lines on paper. The installation emphasizes that a learned model can yield an aesthetic object with modest means—accessible to a broad audience yet grounded in a precise computational procedure. Further documentation and process notes are available on the project site: https://facells.com/.

### A. *Reading what's on the wall.*

Gallery labels use a compact syntax to make the method legible without revealing the whole pipeline. For example, "FaCell 4—Eyeglasses th8.0 maxl2500 s0" refers to the FaCell built from Dense output cell 4 (the CelebA attribute Eyeglasses). "th8.0" is the per-point threshold applied at the Dense output for that attribute; "maxl2500" limits the number of line segments shown; and "s0" is the random seed used for selection and ordering. This notation ties what viewers see to the exact parameters that generated it.

### B. *Public participation and distribution.*

Letter-size FaCells portraits are offered to visitors during the show. They are not free; instead, the "price" is a public social-media post using the hashtags #FaCells and #SpectrumMiami. The exchange is intentional: each drawing leaves the booth as a physical keepsake while simultaneously re-entering the digital stream through the visitor's post. The piece thus circulates in both worlds—affordable to produce and acquire, yet tangible and unique in each impression.

### C. *Conceptual frame.*

This exhibition proposes that machine learning can produce artworks that are at once rigorous and inexpensive to realize. A FaCell is meant to travel easily—plotter lines on letter paper—while remaining an object to keep, a small cult artifact that signals an early instance of AI operating as an artist in a physical medium. The work's route—data → model → line → paper → social post—highlights how computation and culture can loop through one another. As an example of the wall presentation, Figure 7 shows *FaCell 4—Eyeglasses th8.0 maxl2500 s0*.

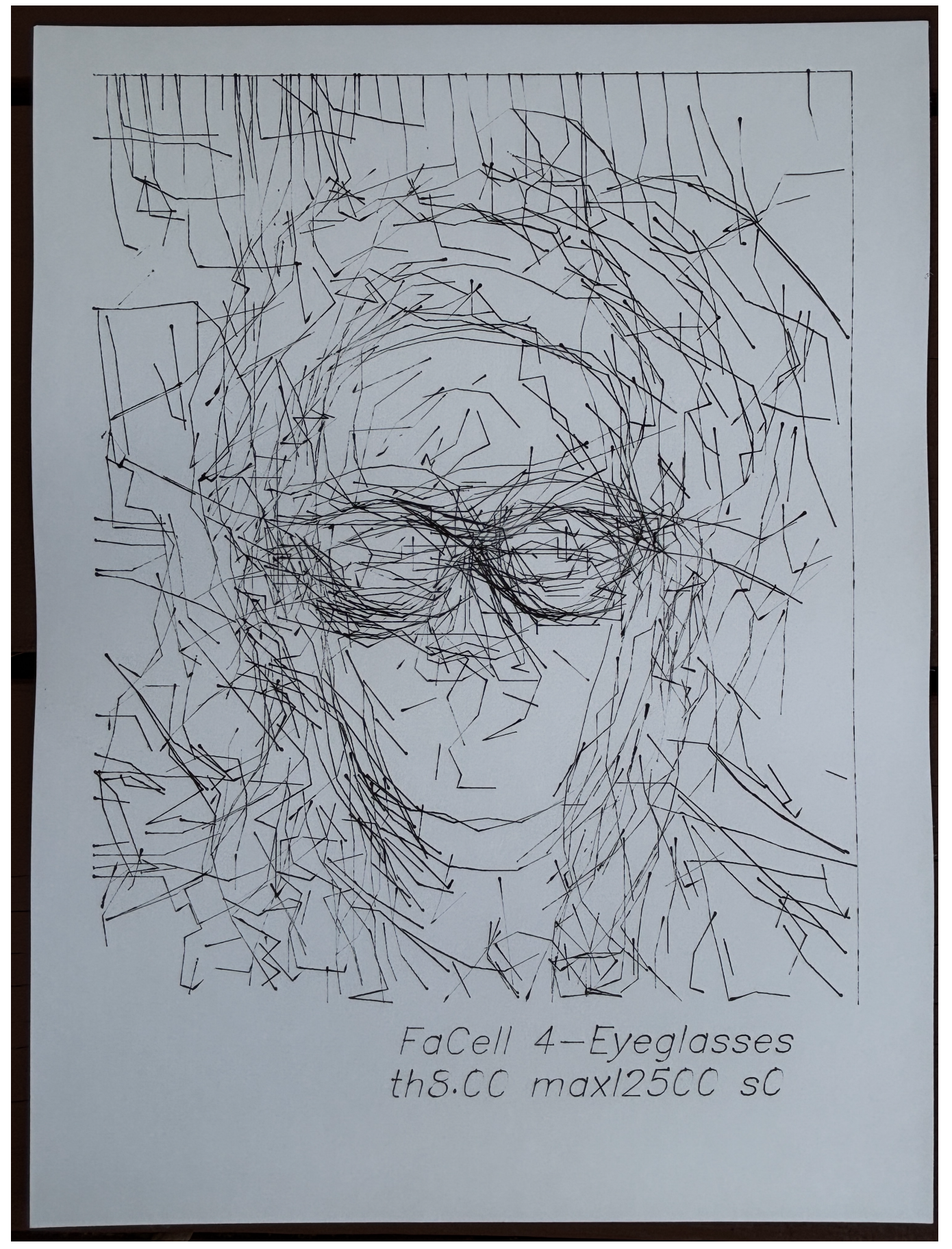

*Figure 7.* ***FaCell 4—Eyeglasses th8.0 maxl2500 s0.*** *Aggregation of per-point selections from many aligned portraits using Dense output cell 4 (Eyeglasses). Threshold 8.0 on the per-point scores; maximum 2,500 lines; seed 0. Plotter drawing on letter-size paper.*

## VII. Discussion, Future Work, and Conclusion

### A. *Discussion—limits, biases, ethics.*

FaCells relies on labels that are convenient but not definitive. CelebA's binary "Male" label simplifies a non-binary reality, and its attributes reflect a particular cultural and temporal moment. The dataset itself—celebrity images with makeup, lighting, and accessories—can imprint correlations that a model may treat as causes. Our per-point values are **model-internal evidence**, not ground truth; they indicate what the network finds diagnostic, not where an attribute inherently "is." When we call the resulting drawings "objective," we mean only that individual human style is minimized: what remains is the projection of many labeled examples through a fixed procedure. Seen this way, FaCells show, in practice, that formal choices in data and models (encodings, ordering, thresholds) generate the meanings we see—and that the resulting aesthetics, in turn, reveal the assumptions built into the pipeline.

### B. *Future work (brief).*

Three paths suggest themselves: (1) stronger vector pipelines—learned strokes and differentiable rasterization; (2) alternative sequence backbones (e.g., transformer-based) and calibrated per-point attributions; and (3) broader, consented datasets and interactive tools that let viewers explore thresholds and seeds directly.

### C. *Conclusion.*

This paper repurposes a classifier's internal computation to **compose** new line drawings: a small architectural change yields per-point attribute scores that, aggregated across many faces, become FaCells. The plotter bridges code and paper, making a process that is rigorous and reproducible also tangible and affordable. In the context of Spectrum Miami 2025, FaCells proposes that machine learning can be read not only as a predictor but as a maker—producing images that move between data and drawing, studio and gallery, digital stream and physical page.

## ACKNOWLEDGMENT

This research was partially supported by Peruilh Ph.D. Scholarship of the School of Engineering of the University of Buenos Aires. The models were trained on the Habanero cluster from Columbia University High Performance Computing (HPC) services.

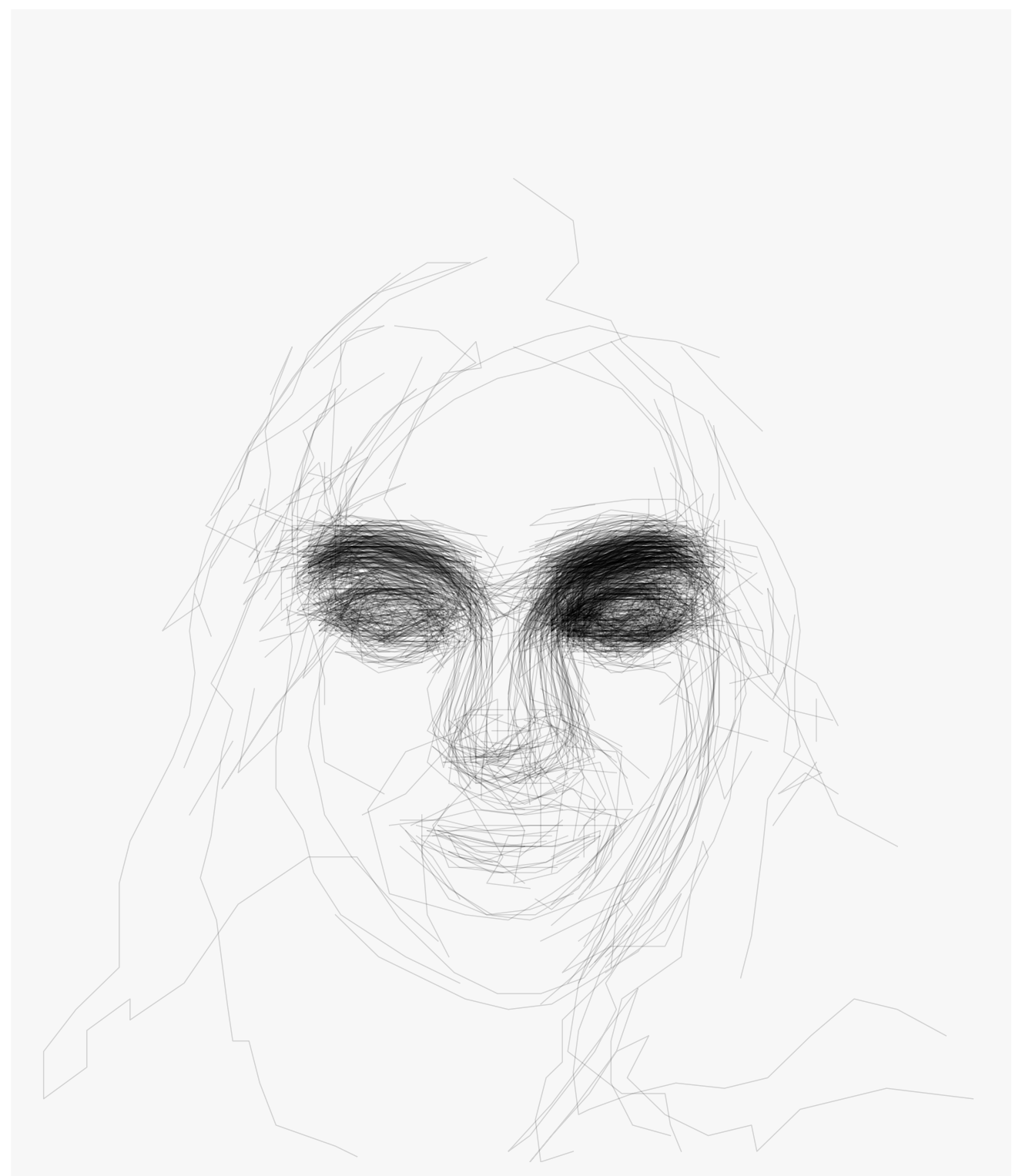

*Arched eyebrows 1000-19*

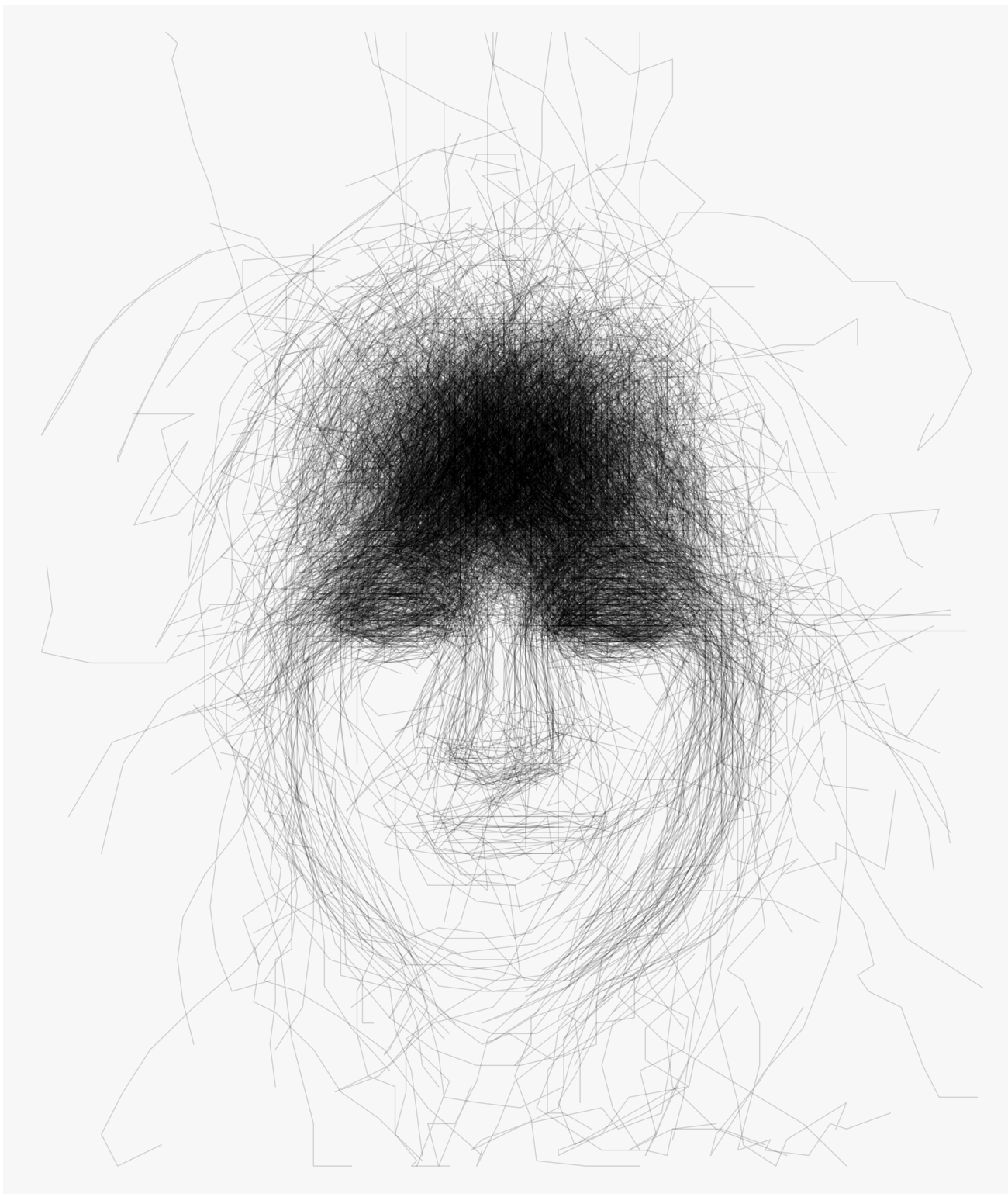

*Bangs 1000-28*

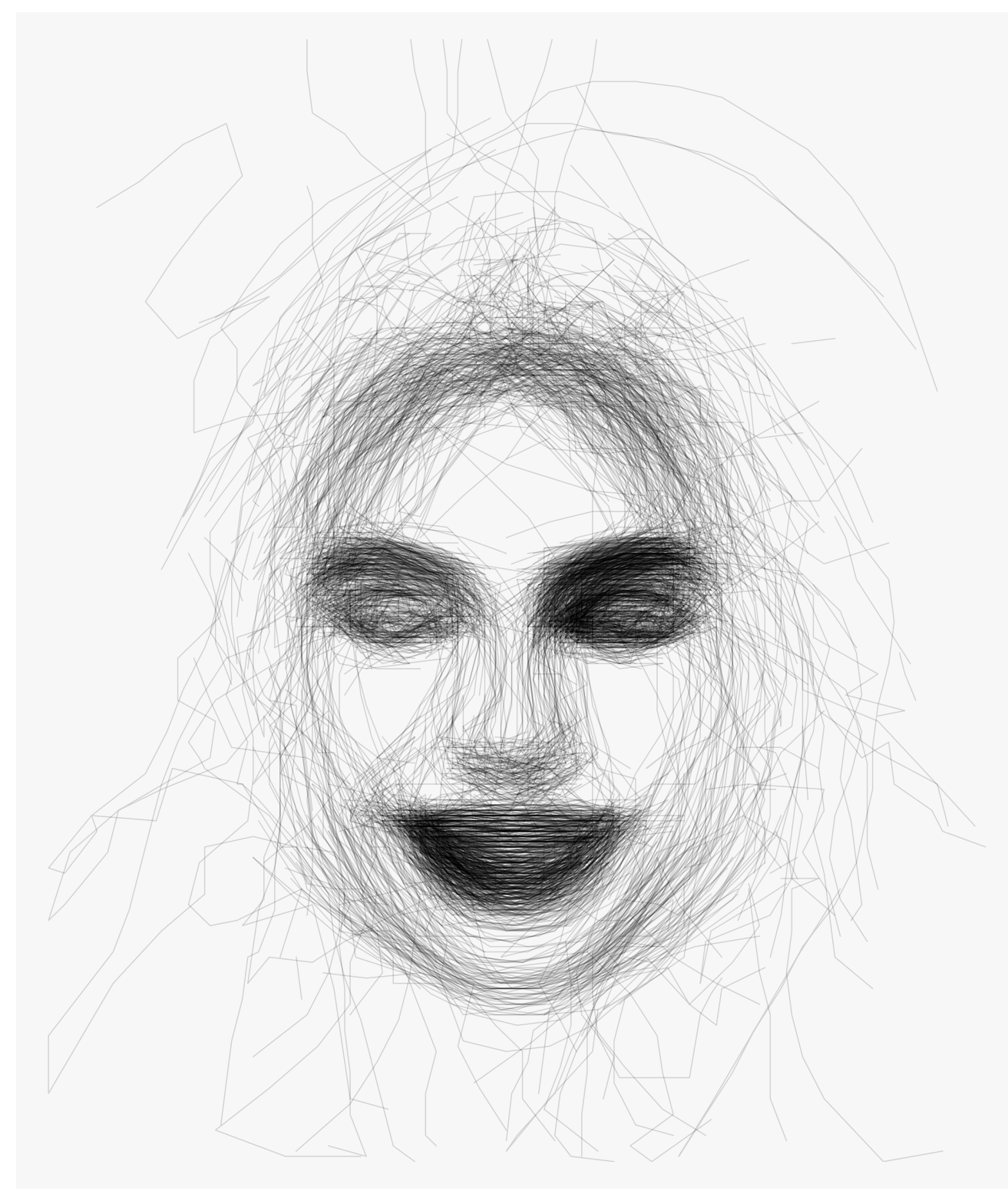

*Attractive 1000-9*

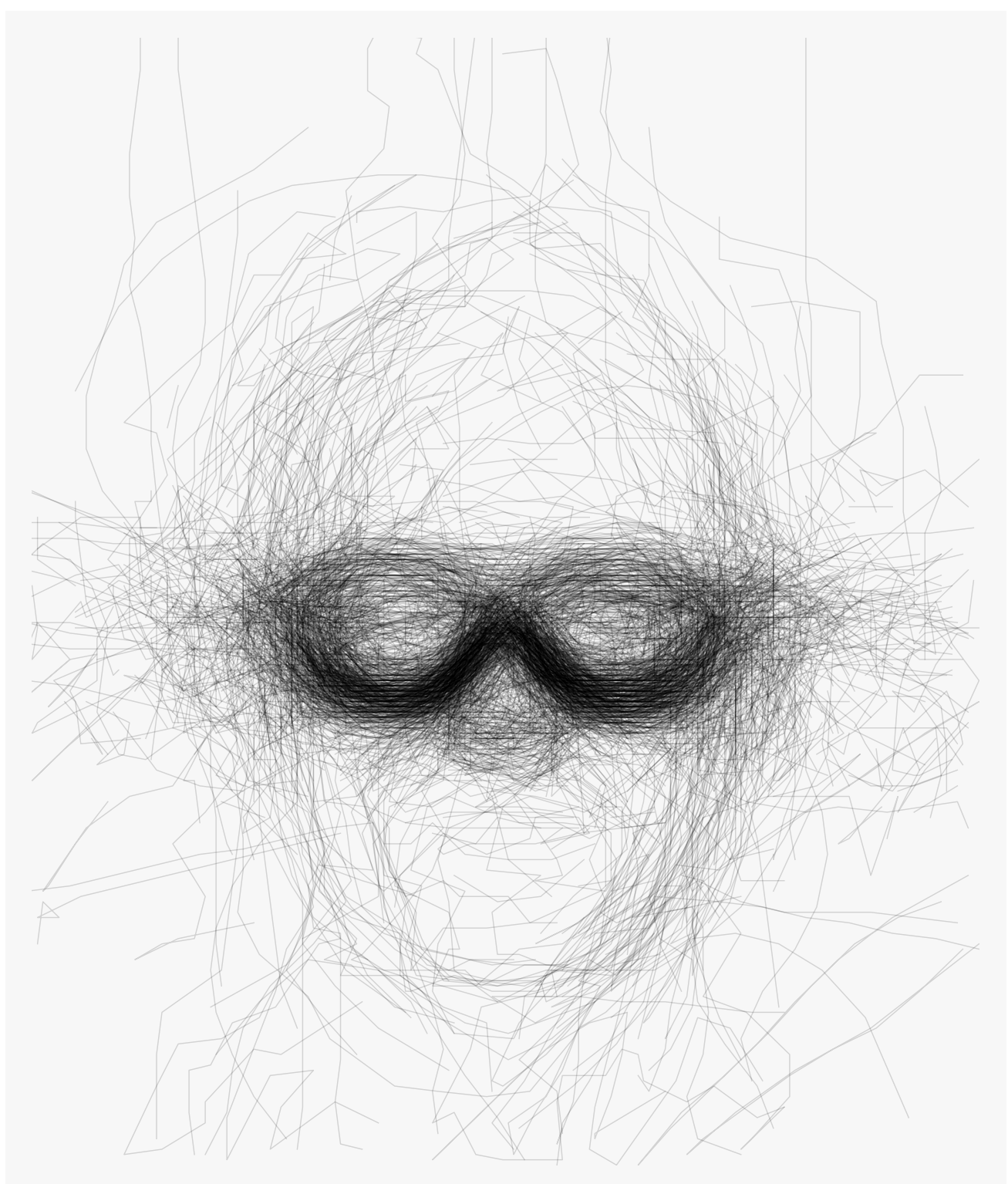

*Eyeglasses 1000-35*

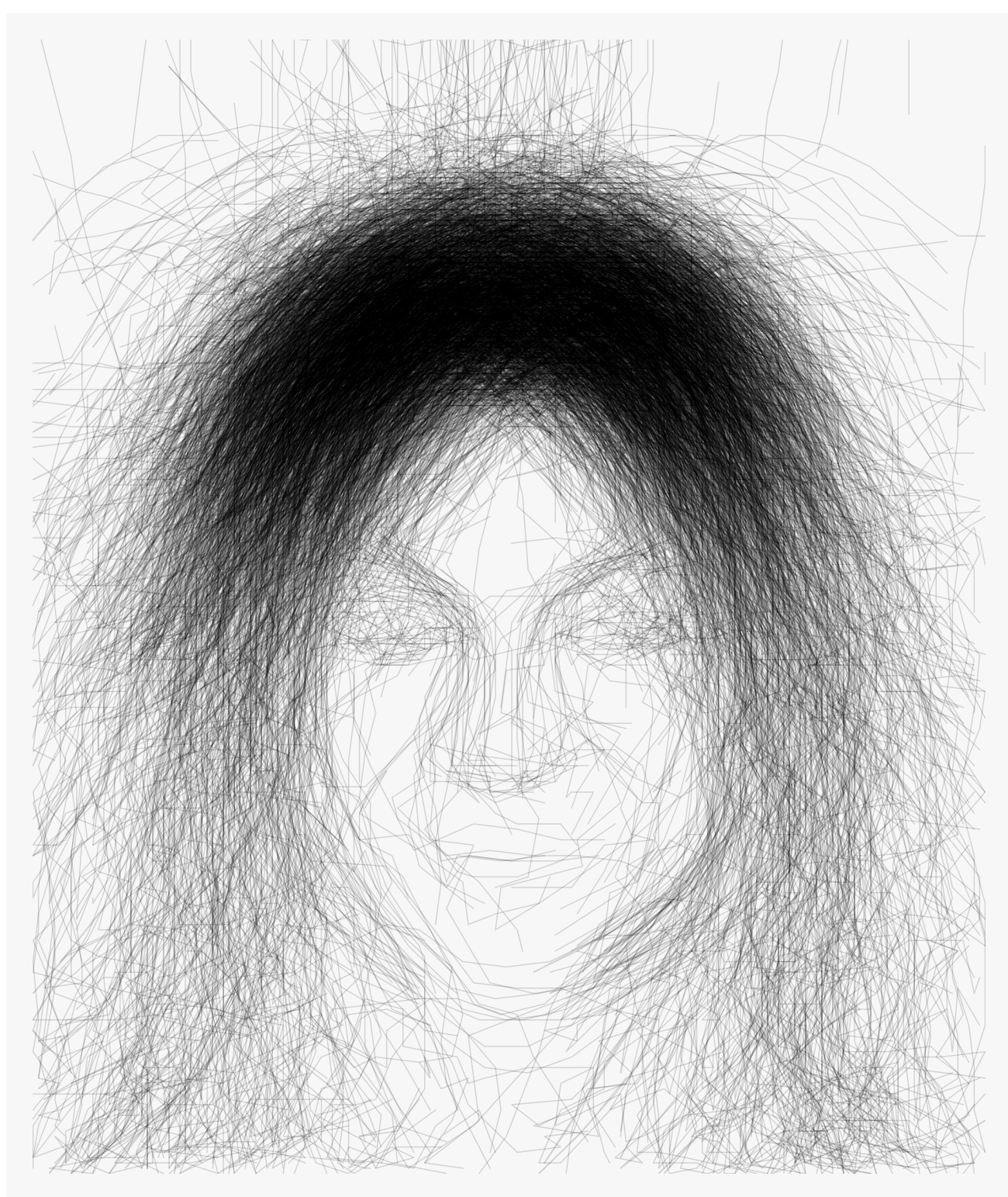

*Blond hair 1000-13*

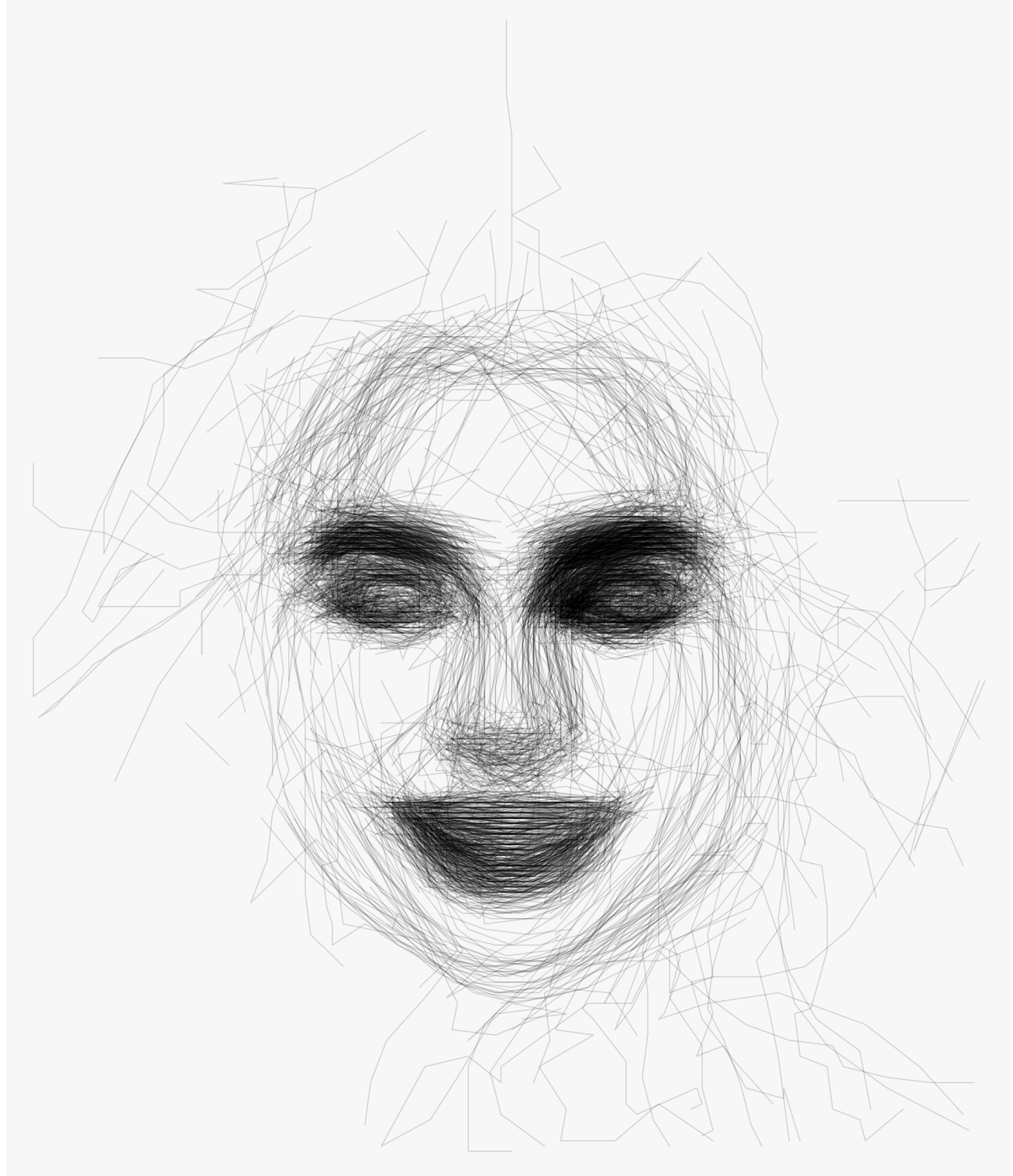

*Heavy makeup 1000-24*

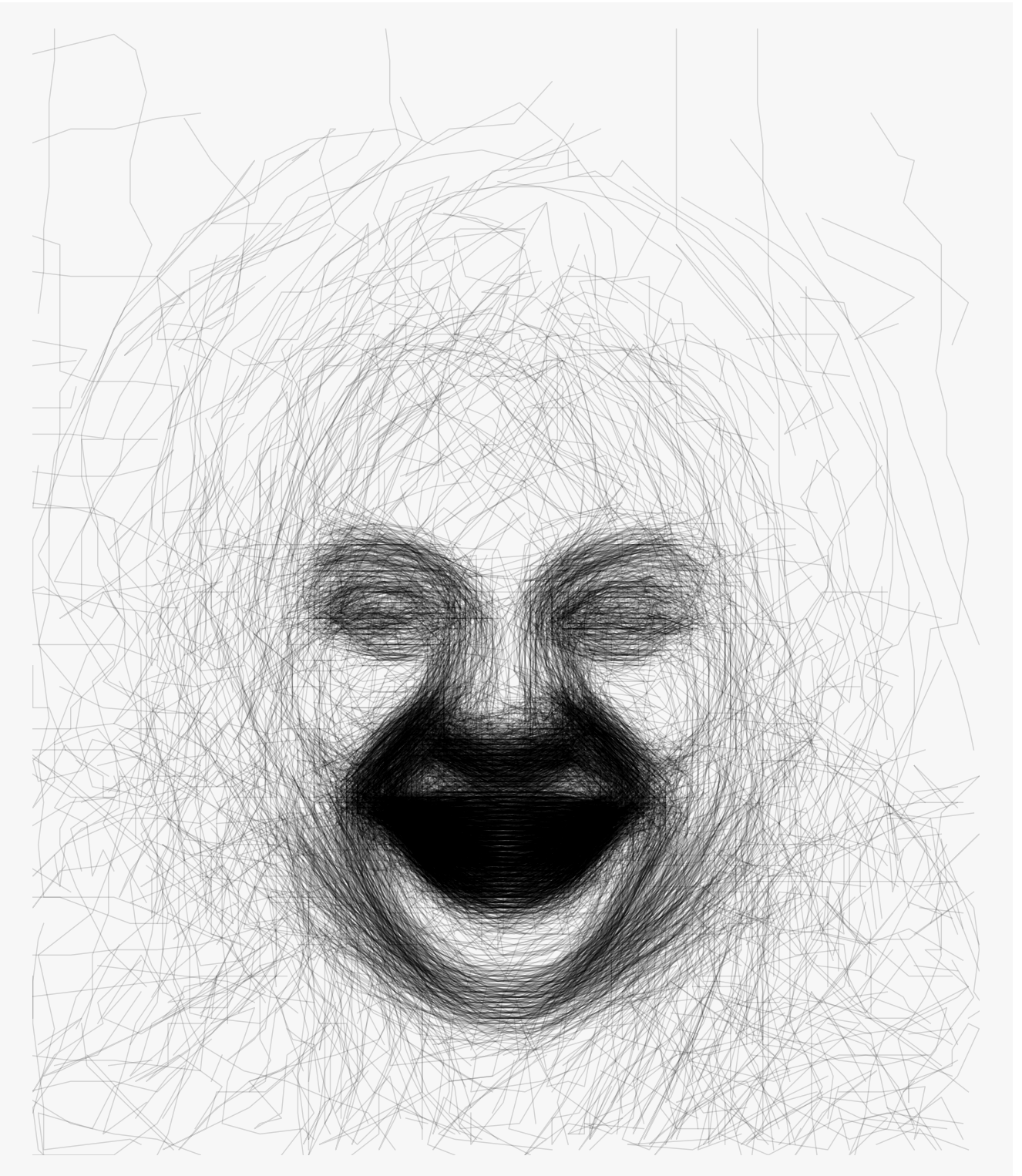

*High cheekbones 1000-16*

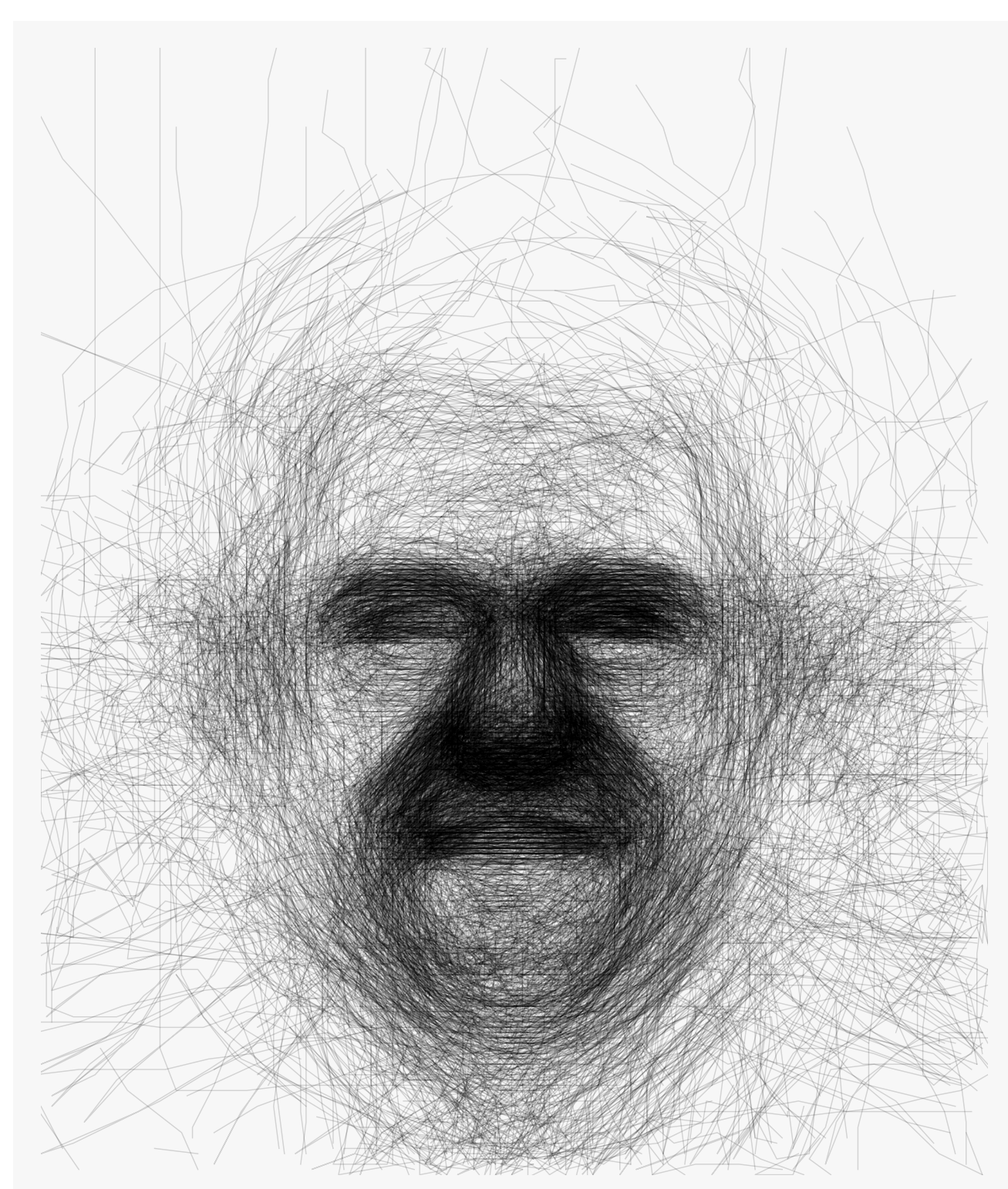

*Male 1000-14*

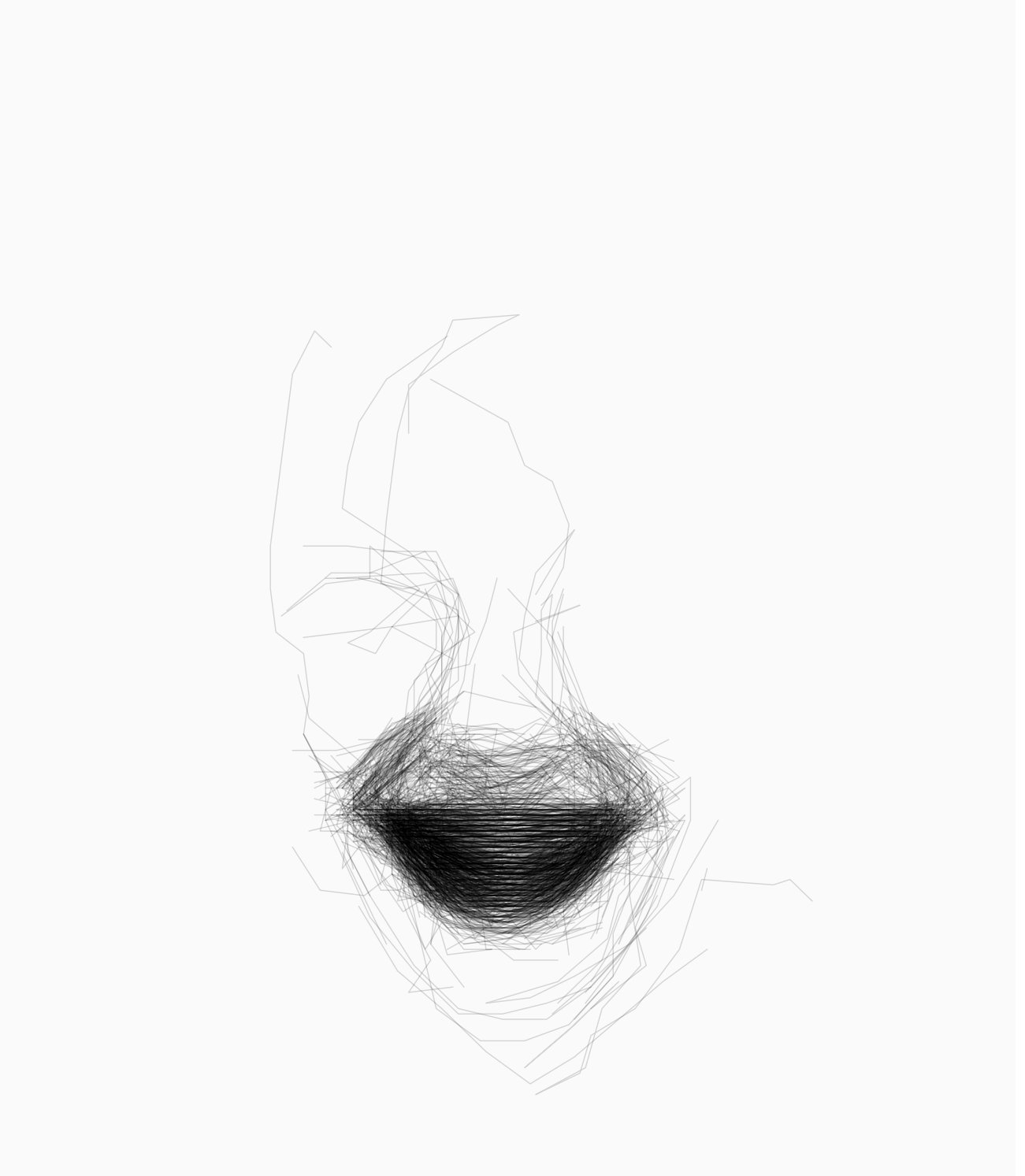

*Mouth slightly open 1000-27*

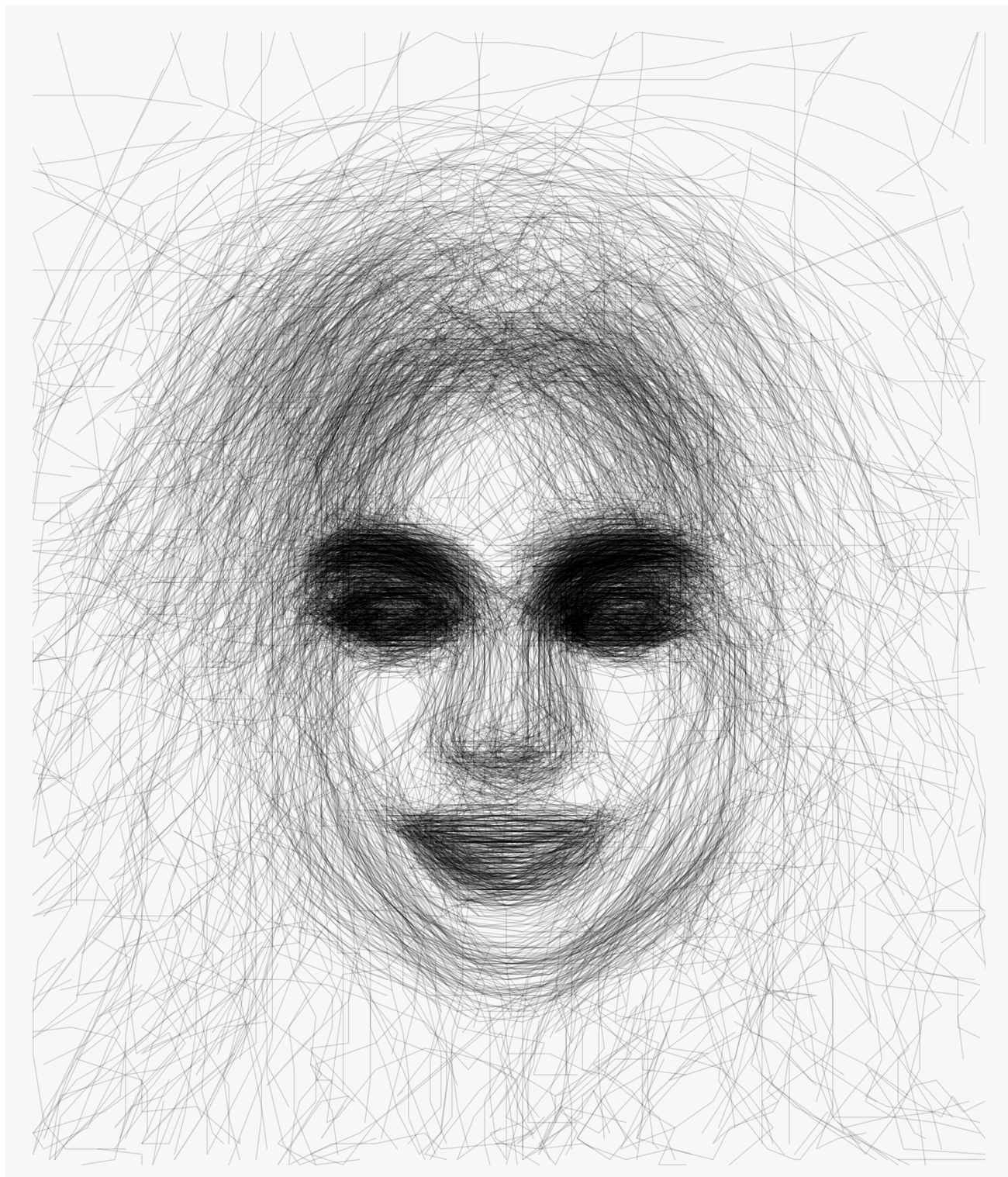

*No beard 1000-12*

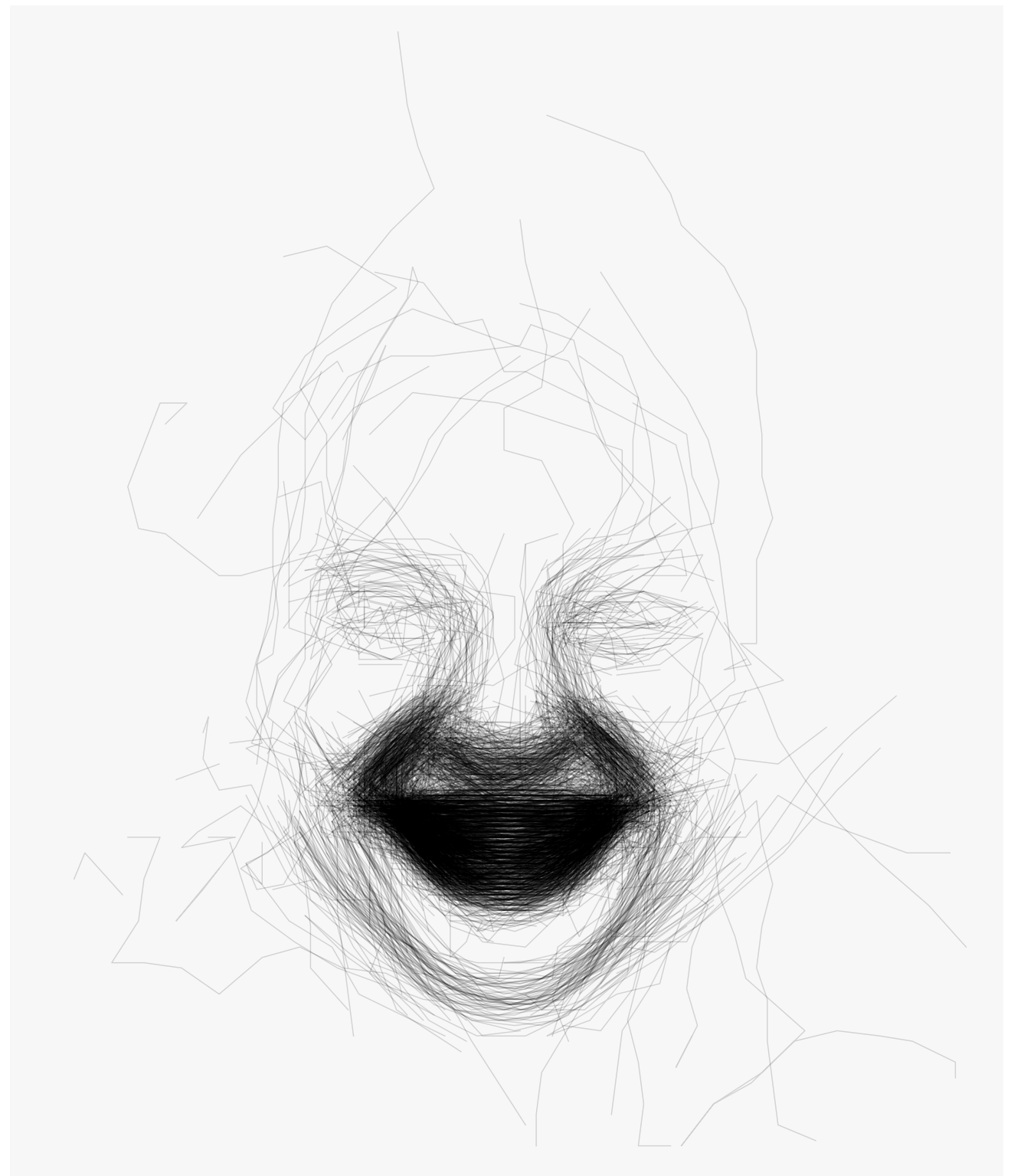

*Smiling 1000-33*

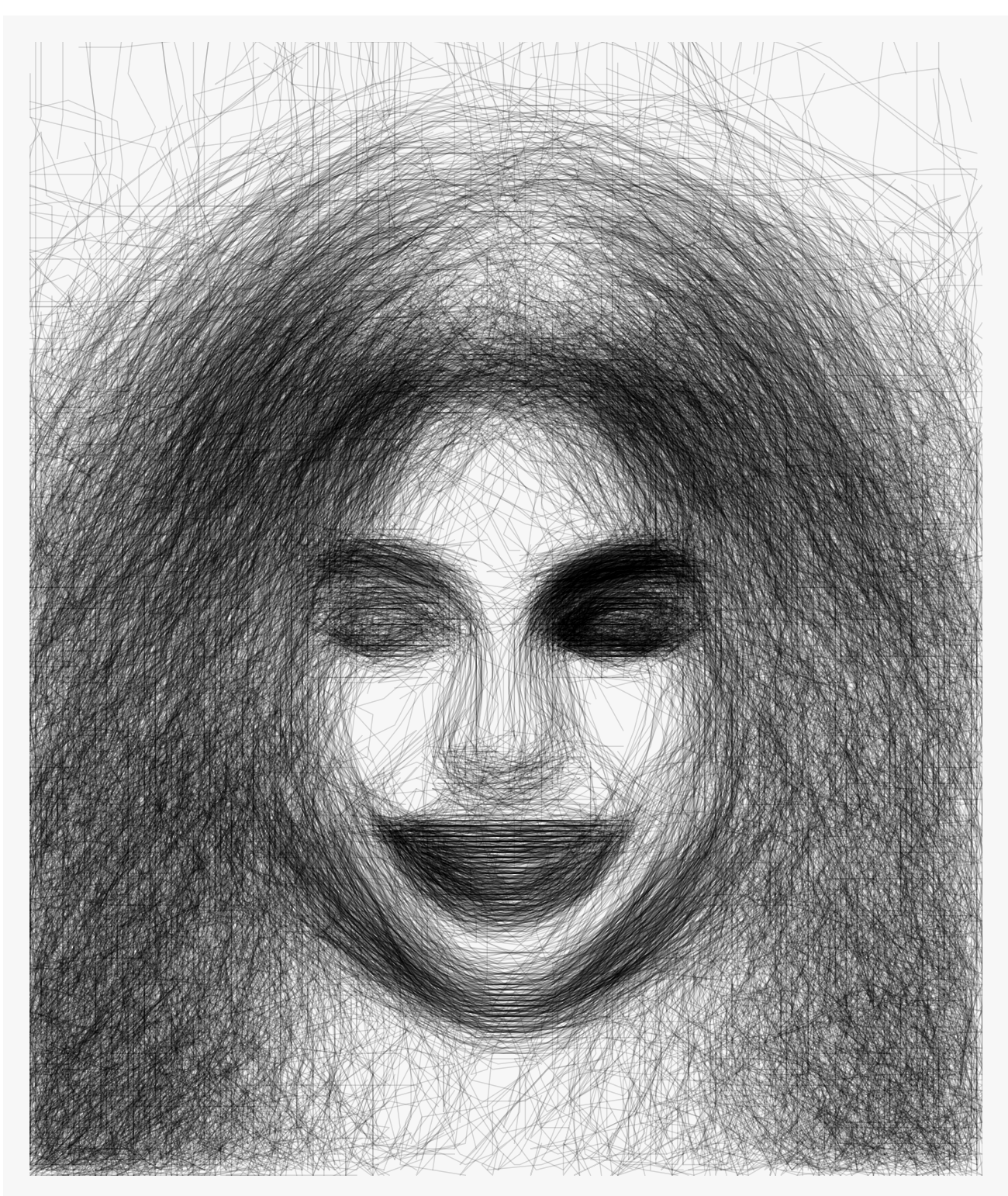

*Wavy hair 1000-4*

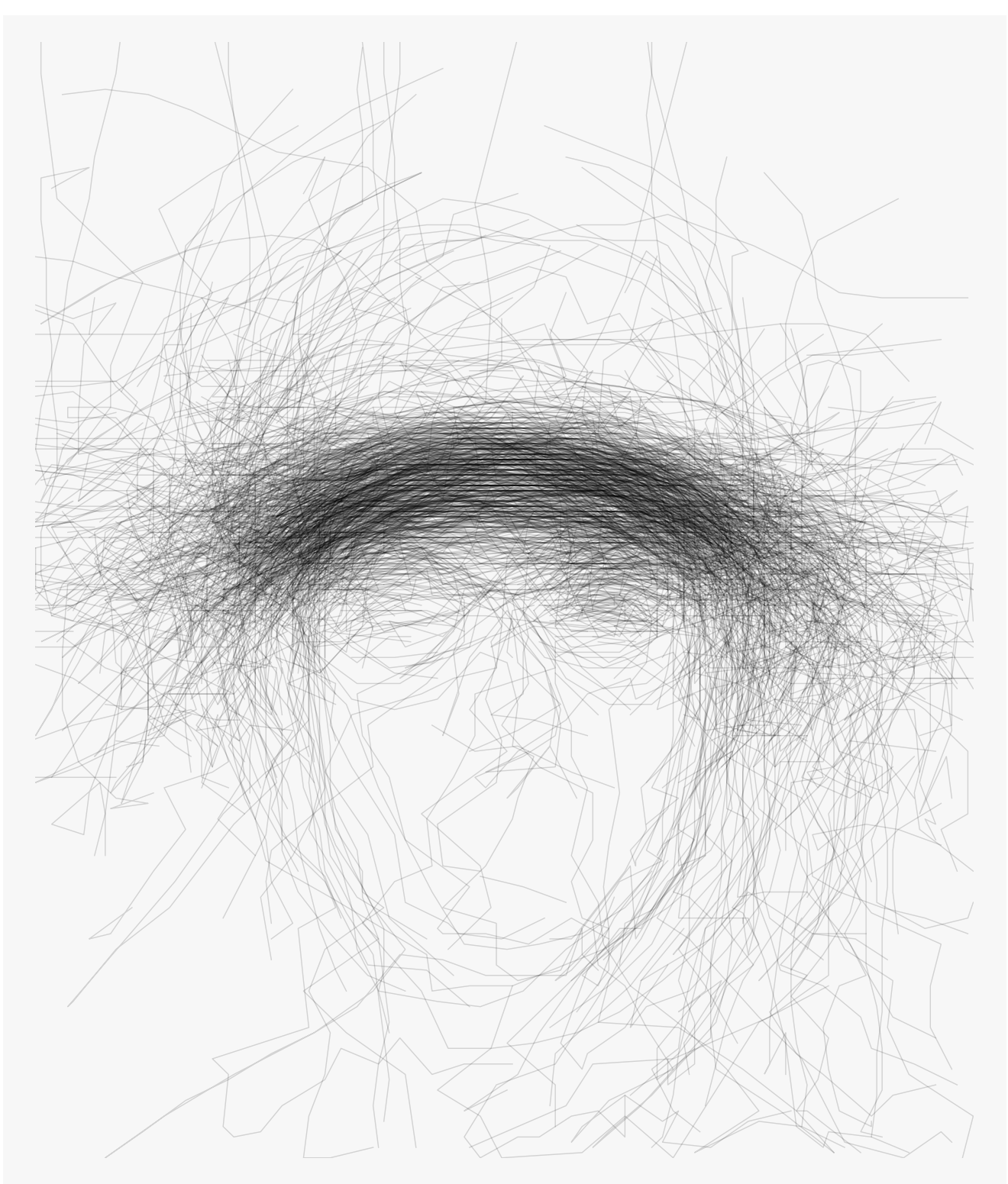

*Wearing hat 1000-35*

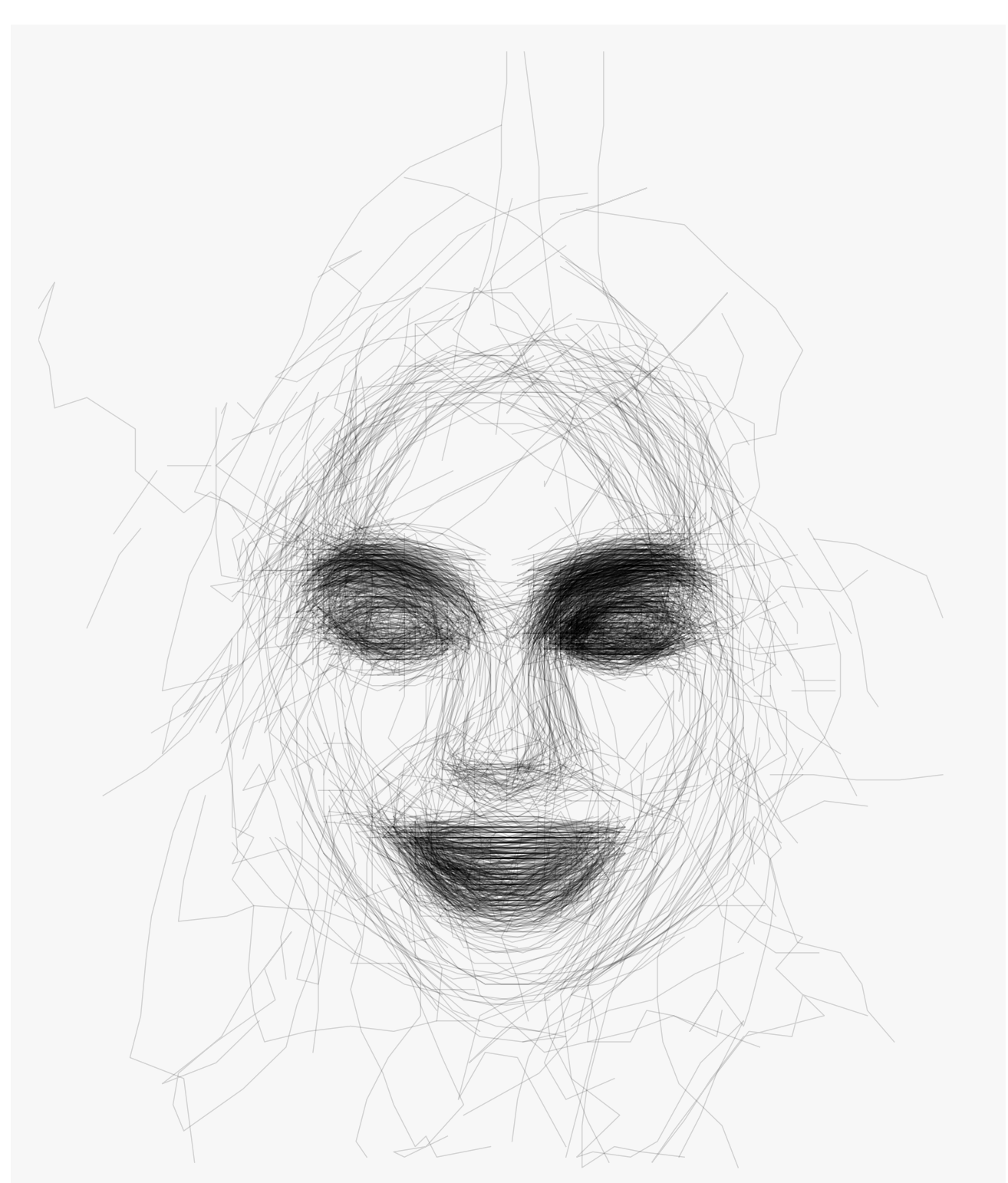

*Wearing lipstick 1000-25*

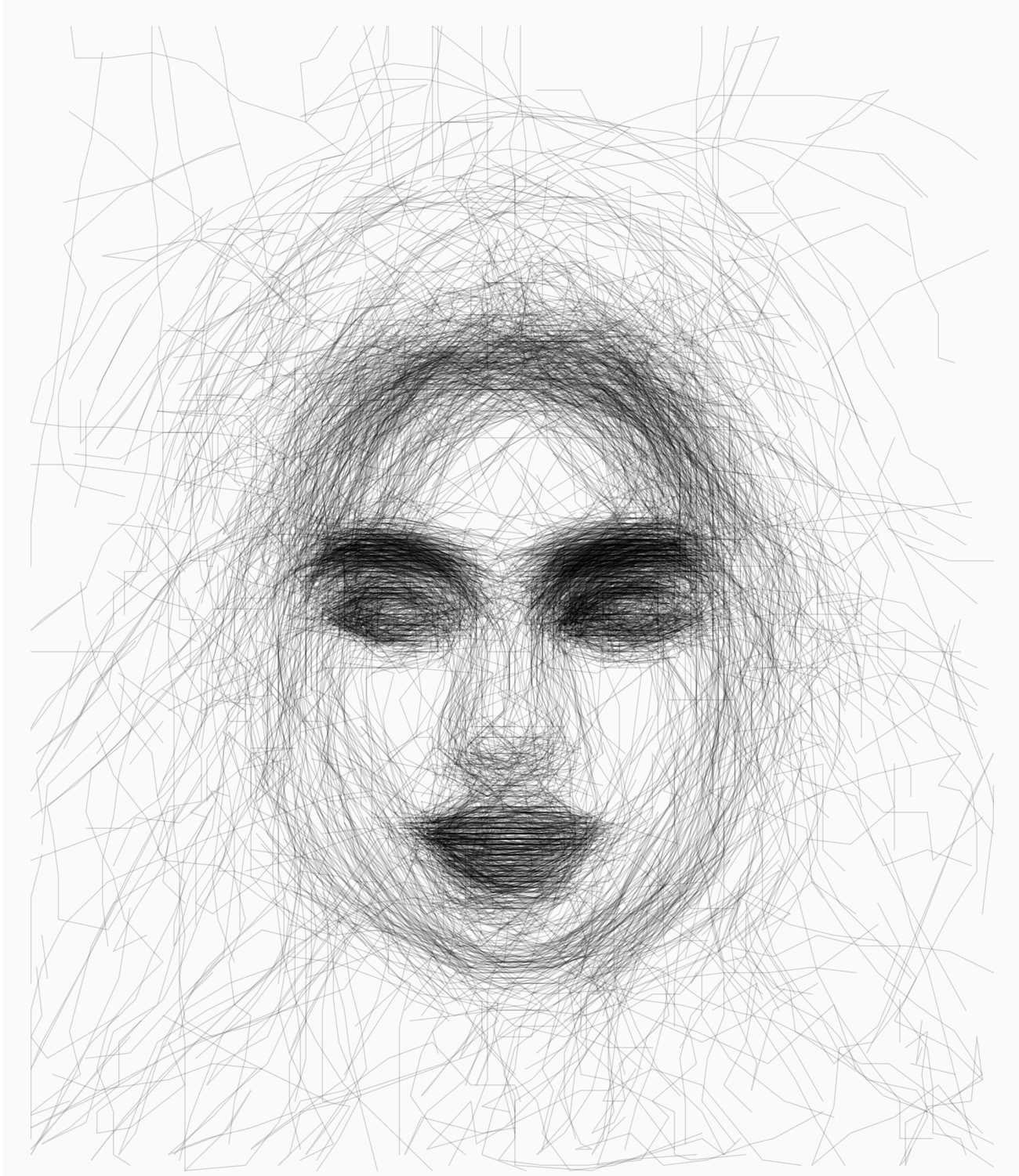

*Young 1000-18*

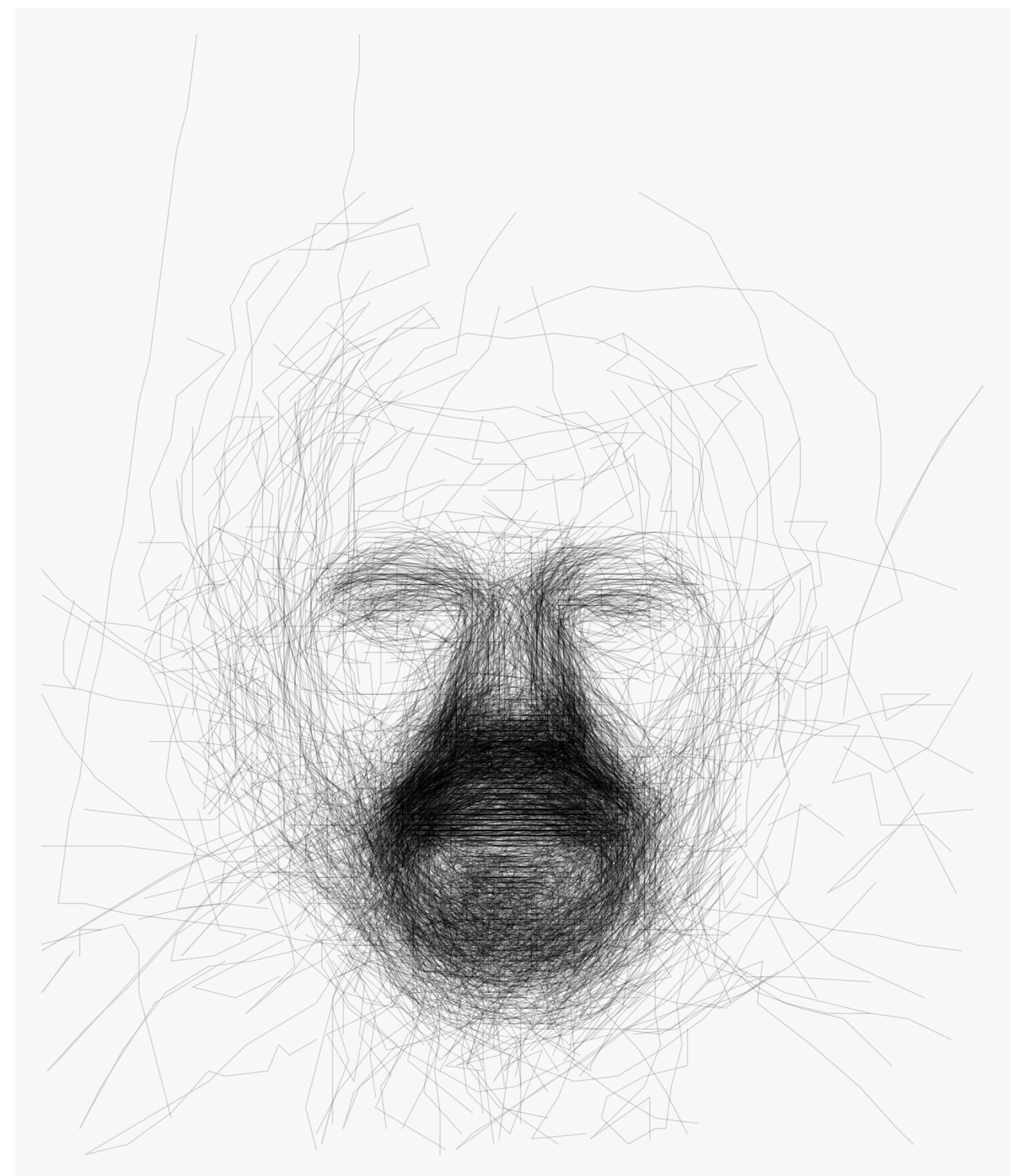

*Not No beard 1000-(24)*

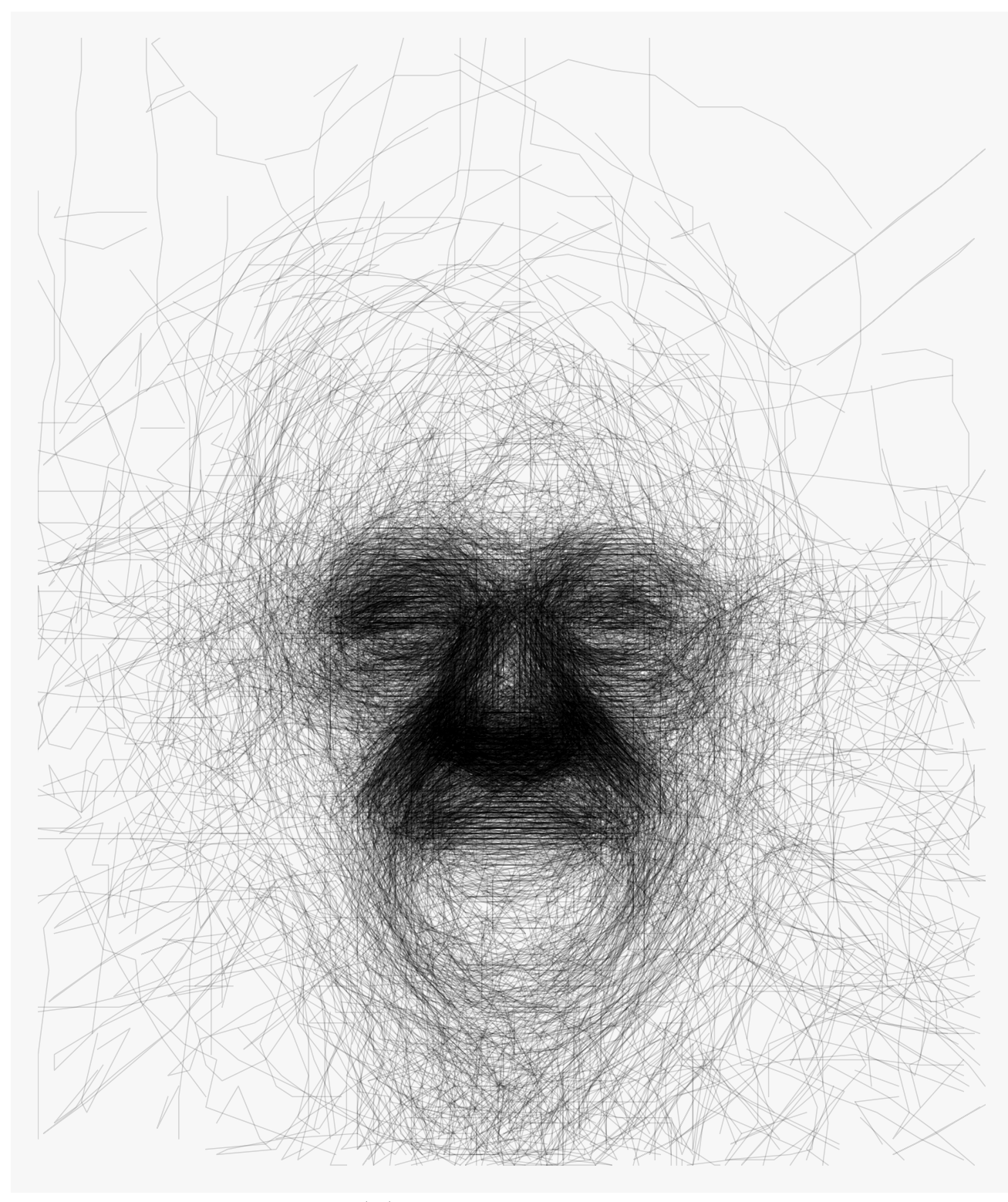

*Not Attractive 1000-(9)*

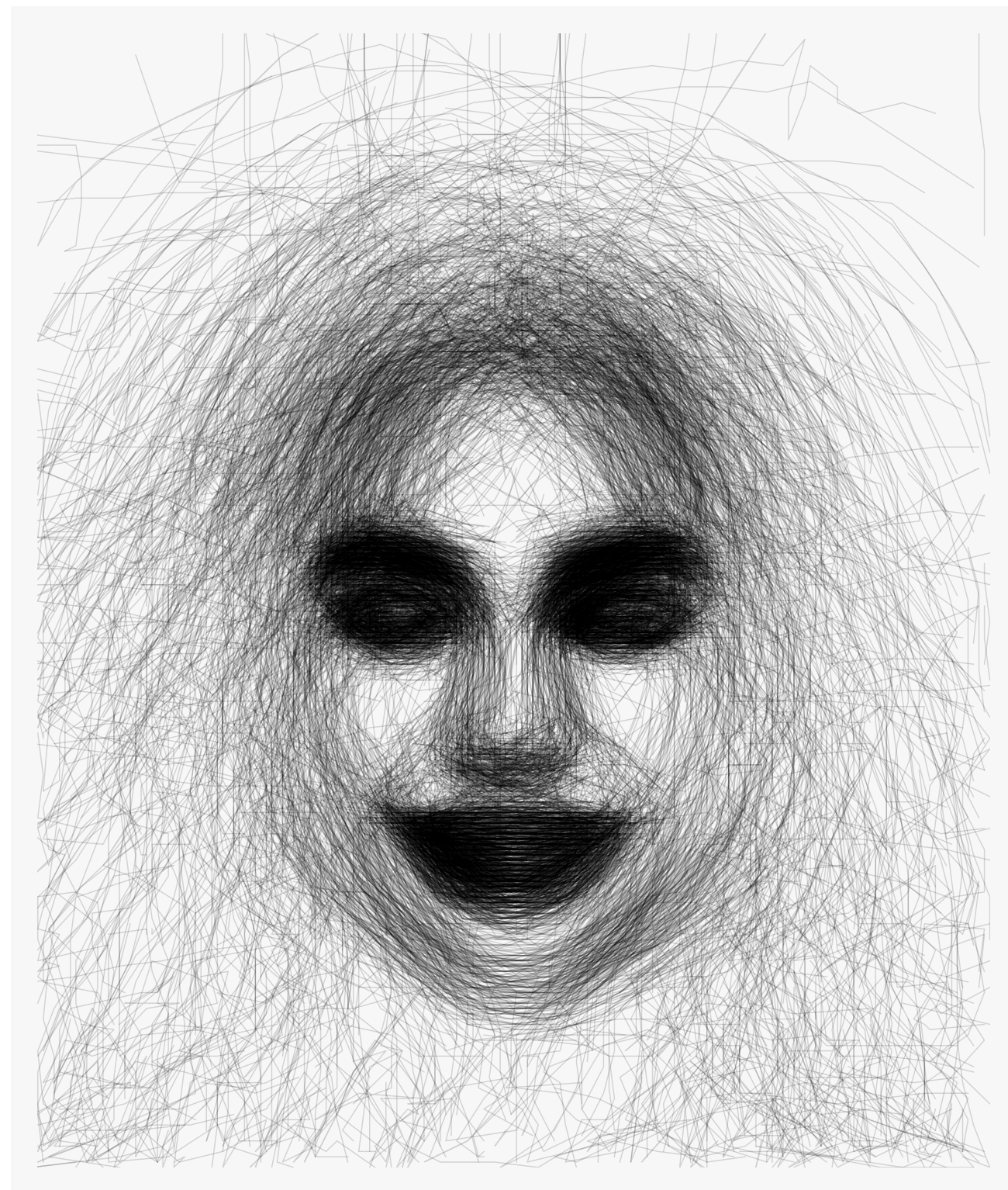

*Not Male 1000-(14)*